\documentclass{elsarticle}
\usepackage{graphicx} % Required for inserting images
\usepackage{orcidlink}

\usepackage{float}
\usepackage{subcaption}
\usepackage{multirow}
\usepackage{makecell}
\usepackage{xcolor}

\newcommand{\new}[1]{#1}

\usepackage{soul}
\usepackage{enumitem}

\begin{document}
\begin{frontmatter}

\title{DeepArUco++: improved detection of square fiducial markers in challenging lighting conditions}

% Authors

\author[1]{Rafael Berral-Soler\corref{cor} \orcidlink{0000-0002-9105-1481}}
\ead{rberral@uco.es}
\author[1,2]{Rafael Mu\~noz-Salinas\corref{con} \orcidlink{0000-0002-8773-8571}}
\ead{rmsalinas@uco.es}
\author[1,2]{Rafael Medina-Carnicer\corref{con} \orcidlink{0000-0003-4481-0614}}
\ead{rmedina@uco.es}
\author[1,2]{Manuel J. Mar\'in-Jim\'enez\corref{con} \orcidlink{0000-0001-9294-6714}}
\ead{mjmarin@uco.es}

% Corresponding/contributors

\cortext[cor]{Corresponding author}
\cortext[con]{Contributing authors}

% Addresses

\address[1]{Dpt. Computing and Numerical Analysis, University of C\'ordoba, Spain}
\address[2]{Maimonides Institute for Biomedical Research of C\'ordoba (IMIBIC), Spain}

% Date

\date{\today}

% Abstract

\begin{abstract}
Fiducial markers are a computer vision tool used for object pose estimation and detection. These markers are highly useful in fields such as industry, medicine and logistics. However, optimal lighting conditions are not always available, and other factors such as blur or sensor noise can affect image quality. Classical computer vision techniques that precisely locate and decode fiducial markers often fail under difficult illumination conditions (e.g. extreme variations of lighting within the same frame). Hence, we propose DeepArUco++, a deep learning-based framework that leverages the robustness of Convolutional Neural Networks to perform marker detection and decoding in challenging lighting conditions.  The framework is based on a pipeline using different Neural Network models at each step, namely marker detection, corner refinement and marker decoding. Additionally, we propose a simple method for generating synthetic data for training the different models that compose the proposed pipeline, and we present a second, real-life dataset of ArUco markers in challenging lighting conditions used to evaluate our system. The developed method outperforms other state-of-the-art methods in such tasks and remains competitive even when testing on the datasets used to develop those methods. Code available in GitHub: \url{https://github.com/AVAuco/deeparuco/}
\end{abstract}

\end{frontmatter}

% Text body

\section{Introduction}
\label{sec:intro}

\begin{figure}[t]
    \centering
    \includegraphics[width=\linewidth]{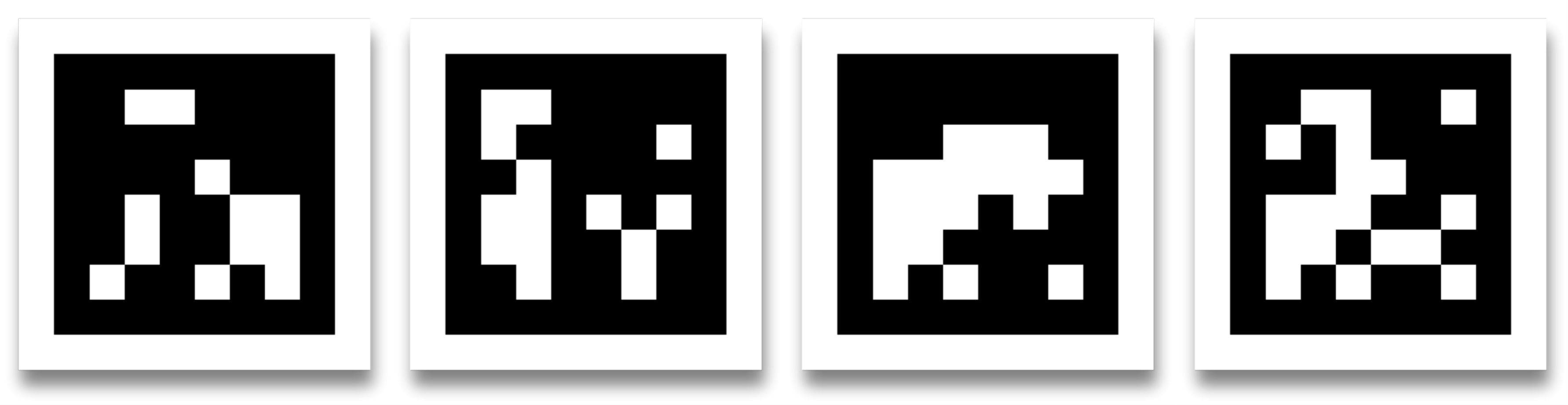}
    %\vspace{0.5mm}
    \caption{\textbf{Some ArUco marker examples.} The type of fiducial marker used in this work. The bits can be read from left to right, top to bottom (black means 0, white means 1), discarding the outer white and black border (the two outermost ``rings'') to produce a sequence of $36$ bits ($6 \times 6$), that is compared against the ArUco dictionary to obtain the encoded ID. (Best viewed in digital format)}
    \label{fig:marker_example}
\end{figure}

Fiducial square planar markers are artificial images used in computer vision applications to facilitate tasks such as camera and object pose estimation, object detection and tracking, and navigation. These markers encode data that allows for precise identification within a scene and have a regular shape that helps to determine their orientation relative to the camera. Examples of such markers are ArUco \cite{garrido-jurado_automatic_2014} (see Fig.~\ref{fig:marker_example}), AprilTag~\cite{olson_apriltag_2011} or ARTag \cite{fiala_designing_2010}, which are utilized on fields such as healthcare \cite{sarmadi_3d_2019,sarmadi_joint_2021} and robotics~\cite{sani_automatic_2017,strisciuglio2018}.

Fiducial markers can also be used as a source of information for 3D reconstruction \cite{3dreconstruction} and mapping techniques like SLAM \cite{ucoslam} (Simultaneous Localization and Mapping). These methods rely on locating reference points (dubbed keypoints) to keep track of an agent's position within a scene. However, finding such keypoints can be challenging, especially in regions that lack texture information. In these cases, fiducial markers can complement existing keypoints or even replace them entirely when valid keypoints cannot be found.

%While useful in controlled environments and having a good throughput, classic fiducial marker detection and identification methods are not robust when facing conditions different from expected, as the techniques on which they are based assume adequate lighting or rely on careful fine-tuning to perform at their best 
While classic fiducial marker detection and identification methods perform well in controlled settings, they are not robust under unexpected conditions, as they require adequate lighting or careful fine-tuning to perform at their best \cite{zhang2023deeptag}. For instance, classic ArUco relies on image contour thresholding and polygon extraction to detect markers in an input image \cite{romero2021track, fiducial_markers}. In this case, blur can make borders challenging to detect, and lighting conditions can render thresholding settings useless.

On the other hand, convolutional neural networks (CNNs) have seen great success in tasks such as image object detection and recognition, being more robust and flexible than classic computer vision techniques. Since the success of AlexNet \cite{krizhevsky_imagenet_2012}, deep CNNs have retained their state-of-the-art status in such tasks without additional restrictions. These methods can improve robustness towards blur and lighting through the use of data augmentation techniques \cite{dataaugmentation}.

\begin{figure}[t]
    \centering
    \includegraphics[width=\linewidth]{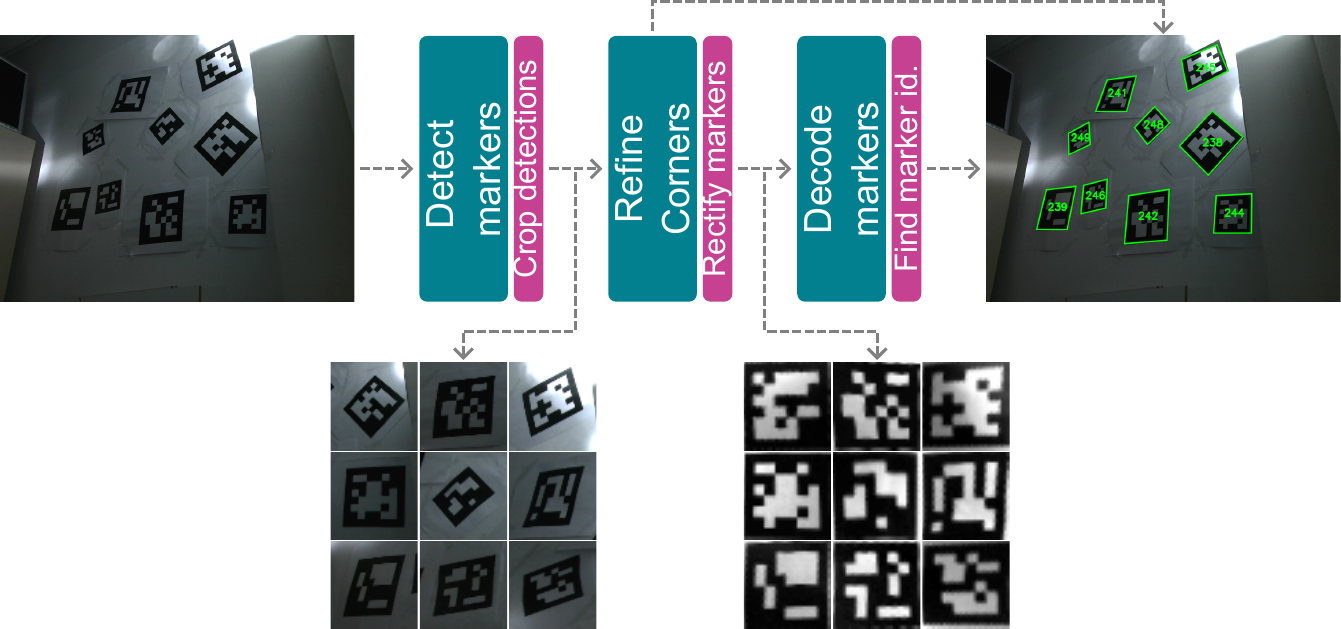}
    \vspace{0.5mm}
    \caption{\textbf{Main: DeepArUco++ framework.} The marker detector receives as input a color image and returns a series of bounding boxes, which are used to obtain crops from the input image. Then, the corner regressor model is applied over each crop to refine the position of the corners. The detected markers are then rectified with the refined corners and used as input for the marker decoder. Finally, the marker is assigned the ID with the least Hamming distance w.r.t. the decoded bits. (Best viewed in digital format)}
    \label{fig:full_pipeline}
\end{figure}

Inspired by this, we propose a novel method (Fig.~\ref{fig:full_pipeline}) using deep CNNs to perform the detection, precise location, and decoding of square fiducial planar markers. While we target ArUco markers, the method should readily adapt to other square planar markers, provided they encode information in a similar way. We needed a large dataset of ArUco markers in varying orientations and lighting conditions to develop our methods. As we did not find any publicly available datasets meeting our requirements, we developed a method for creating a synthetic dataset that can be used to train our models, providing a high degree of flexibility in the generation of samples. Additionally, we captured a real-life dataset of ArUco markers in challenging lighting conditions to test our approach and provide a fair comparison with preexisting methods.

% contributions
The contributions of this work are the following: 
\begin{itemize}[noitemsep,nolistsep]
    \item First, we present a new framework composed of three separate models: marker detector, corner regressor, and marker decoder, based on deep CNNs. These models are combined in a pipeline to perform detection, precise location, and decoding of ArUco markers. Using disjointed models at each step allows for a more straightforward approach to the problem and favors modularity. This method outperforms classic ArUco and DeepTag~\cite{zhang2023deeptag} in challenging lighting conditions while keeping a competitive throughput. Also, the presented method stays competitive compared to other datasets (namely the DeepTag dataset).
    \item Second, we introduce a simple but effective method for generating synthetic datasets containing ArUco markers, used for the training of each model in the pipeline. 
    \item Third, we captured a second, real-life dataset of ArUco markers in challenging lighting conditions for testing our methods. Finally, the implementation of our methods and the datasets we produced will be published online for public use (see the URL provided in the Abstract). 
\end{itemize}

While our method incorporates knowledge from previous works in object and keypoint detection, our proposed combination, trained on a specially composed synthetic dataset designed to replicate challenging lighting and shadow effects, results in highly robust marker detection even under harsh lighting conditions. To the best of our knowledge, this is the first work that aims to achieve precise marker detection and decoding under such difficult lighting conditions.

An earlier version of this work was presented at IbPRIA'23~\cite{berral2023ibpria}. The main improvements in this new work are the following: first, we aim to improve the detection capabilities of our method and explore a new method for marker corner location based on the use of heatmaps; additionally, we compare our models against other state-of-the-art techniques both on our test dataset and on the DeepTag dataset,  from a performance standpoint and from the perspective of the throughput of our models.

% Rest of paper
The rest of the paper is organized as follows. First, Section~\ref{sec:relworks} summarises some related work. The datasets created for developing and testing our methods are introduced in Section~\ref{sec:datasets}. Then, we describe our proposed approach in Section~\ref{sec:methods}. Section~\ref{sec:results} presents our experiments and discusses their results. Finally, we conclude the paper in Section~\ref{sec:conclusions}.
\section{Related Works} \label{sec:relworks}
Square planar fiducial markers \cite{fiducial_markers} are a popular computer vision tool used whenever precise camera pose estimation is required, used in tasks such as robot navigation and object tracking in augmented reality. %. Some applications of these markers are autonomous robot navigation and object tracking for augmented reality tasks (e.g. using augmented reality to simulate the operation of specialized tools and machinery, often for learning purposes).
ArUco \cite{romero-ramirez_speeded_2018}, AprilTag \cite{wang_apriltag_2016}, and ARTag \cite{fiala_designing_2010} are some of the most popular ones, used in both industry and academia; a previous implementation of ArUco \cite{garrido-jurado_automatic_2014,aruco_dict} is available for its use in the OpenCV computer vision library.

These markers usually follow a similar structure: they are shaped as a square, with black borders that mark the boundaries of a grid of black and white square cells, each representing a different bit value. Information can be encoded in every marker using those bits, allowing for the precise identification of individual markers in the scene. 

\subsection{Classical techniques.}

Fiducial markers are usually detected and decoded through classical computer-vision techniques. These techniques assume adequate lighting and the absence of additional factors impacting image quality and can fail outside controlled conditions. Some works based on classical techniques aim to perform in suboptimal conditions. Authors in \cite{mondejar2018eswa} address the problem of fiducial marker identification under different adverse conditions, such as blur and non-uniform lighting. However, they do not aim to precisely locate the markers in the scene. On the other hand, authors in \cite{romero2021track} locate markers in images achieving resilience towards blurriness, but they do not address the challenge of inadequate lighting. To our knowledge, the task of jointly locating and decoding fiducial markers under challenging lighting conditions remains unaddressed.

Despite the importance of lighting for developing robust object detectors, there is limited literature on object detection under poor lighting settings. Without specialized hardware, such as IR cameras, the most straightforward approach would be to apply a classic image-enhancing technique to achieve better contrast and then use an off-the-shelf object detector over the corrected images. Some of those enhancing techniques are based on thresholding (binary, Otsu \cite{otsu}), gamma correction or histogram equalization. Although straightforward, these techniques often depend on carefully selected values (e.g., gamma or threshold values) and may not account for extreme lighting variations between regions of the same image. Adaptive histogram equalization techniques, like the ones described in \cite{h_eq} do adjust to different regions of the image, but can lead to amplified noise on homogeneous regions, making object detection even more challenging. %Given these limitations, methods relying on these enhancement techniques are not suitable for unattended use when the on-site lighting conditions are not known.

\subsection{Deep Neural Networks.}

%More recently, 
Methods based on deep learning, such as the ones described in \cite{llnet} or \cite{enlightengan}, learn the correspondence between low-lighting images and their normal lighting counterparts, leading to better results than those obtained through the use of classical techniques. However, using a neural network as a preprocessing step is computationally intensive and may lead to longer processing times. This could be mitigated by end-to-end training of a single detector model, starting from in-the-wild images taken under varying lighting conditions or using data augmentation to simulate them, making the image enhancement step unnecessary.

In the past few years, there have been some attempts to tackle the problem of joint marker detection and identification using Convolutional Neural Networks. In \cite{hu2019charuco}, a method for detecting boards of markers under low-light and high-blur conditions is proposed. However, this method is designed around a fixed configuration of markers on a board (referred to as ChArUco) and cannot be used to locate and decode individual markers. Authors in \cite{li2020arucoyolo} leverage YOLOv3 to detect ArUco markers under occlusions. However, this work focuses only on marker detection, not corner refinement or marker identification; lighting conditions are also ignored. Finally, a new work has been published \cite{zhang2023deeptag}, using a MobileNet backbone to detect markers and iteratively refine their corners. However, their work focuses mainly on sensor noise and motion blur; thus, it does not address the problem presented in this work. %Also, they have published the dataset used for developing their methods.

Our proposed approach addresses the problem of detecting markers in challenging lighting conditions in the following manner. First, we developed a new synthetic dataset featuring markers under varying lighting conditions. While mild blur has been used as a data augmentation technique, this is done to obtain a more robust model (i.e. our work does not explicitly tackle blur conditions). Next, we trained a pipeline consisting of an off-the-shelf marker detector along with custom-designed corner refinement and marker decoding models using this dataset. While recently sophisticated methods based on transformers \cite{detr1,detr2,detr3,detr4} have achieved significant success in general-purpose object detection tasks, we have chosen to use YOLOv8 \cite{yolov8} as our object detector due to its efficiency and ease of training and use on mid-range computers. The resulting system can detect and correctly identify ArUco markers under challenging lighting settings, outperforming other state-of-the-art methods (namely ArUco~\cite{romero-ramirez_speeded_2018} and DeepTag~\cite{zhang2023deeptag}) in this task.
\section{Datasets} \label{sec:datasets}

We have used the following datasets to develop and test our methods and compare them against preexisting ones.

\subsection{Flying-ArUco v2 dataset} \label{ssec:flyingaruco}

To train our methods, we have developed a synthetic dataset, dubbed Flying-ArUco v2 dataset, composed of images containing various ArUco markers overlaid on backgrounds sampled from the MS COCO \cite{mscoco} 2017 training dataset. 

%As the initial step for building our dataset, we have sampled $2500$ images from the entire MS COCO 2017 training dataset. 
To create our dataset, we initially sampled $2500$ images from the entire MS COCO 2017 training dataset. These images have been rotated to get a wide aspect ratio and cropped to get valid backgrounds of size $640 \times 360$, discarding images smaller than this target size. \new{We sampled our backgrounds using the following approach: first, we calculated the median \textit{luma}~\cite{luma} (or \textit{lightness} component) for each image by converting it to grayscale using the \texttt{cv2.COLOR\_BGR2GRAY}\footnote{Color conversions on OpenCV: \url{https://docs.opencv.org/3.4/de/d25/imgproc_color_conversions.html}} conversion. This conversion is needed to obtain a single brightness value per pixel, as using the original RGB image (with three values per pixel) would not yield a meaningful result after the median calculation (we would just be computing the most common intensity value for \textit{any} channel). Next, we sorted the images into 10 equal-sized bins, meaning each bin contained the same number of images, although the range of brightness values differed across bins. This binning method allowed us to match the brightness distribution of the source dataset. Finally, we sampled an equal number of backgrounds from each bin to ensure a balanced representation. Overall, the luma range for our samples extends from $0.0039$ to $0.9608$, on a scale of $0$ to $1$.}
%For each image, we computed the median \textit{luma} value (by converting the image to grayscale and then obtaining its median value), and we sorted the pictures in $10$ bins using this value as a criterion. 

Up to $20$ markers sampled from the ArUco DICT\_6X6\_250 dictionary (composed from 250 ArUco markers with $6 \times 6$ bits, for more information please refer to the OpenCV ArUco documentation\footnote{ArUco on OpenCV: \url{https://docs.opencv.org/4.x/de/d67/group__objdetect__aruco.html}}), with varying sizes, orientations, and camera focal distances, have been overlaid on each resulting background. Fake markers have also been included to provide negative examples: full-black markers, markers with inverted colors, markers containing colors different from black or white, and other markers containing patterns composed of random lines and Perlin noise instead of the expected grid of black and white square cells.

\begin{figure}[t]
\includegraphics[width=1.\linewidth]{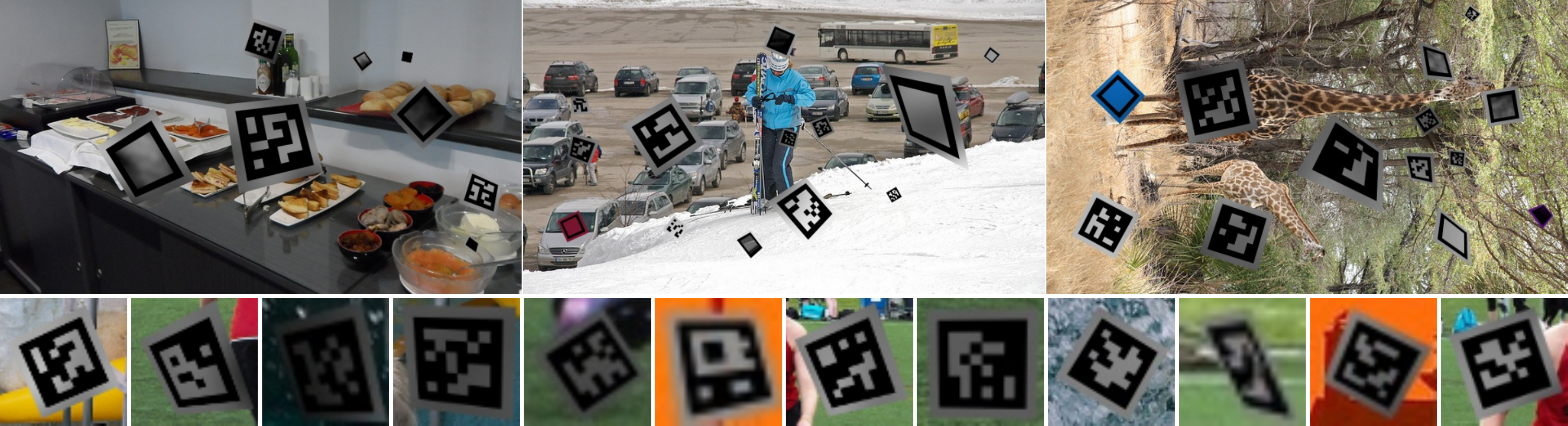}
\vspace{0.5mm}
\caption{\textbf{Flying-ArUco v2 dataset.} Using images from the COCO dataset as background, challenging training samples are created by overlapping markers with varying poses, sizes, and positions in the image. Top: detection dataset. Bottom: refinement/decoding crops. (Best viewed in digital format)}
\label{fig:flyingarucov2}
\end{figure}

Finally, to get a more natural result on the synthetic images, markers have been combined with the background by projecting the luma of the background image over the overlaid markers; doing so makes the lighting of the overlaid markers more consistent with the lighting of the background, making it more challenging to find the added markers just by picking the items with odd lighting. As the background images present a wide variety of lighting settings, the resulting images contain markers under a myriad of lighting and shadow combinations. Please note that this dataset is intended to be used as a tool in the development of marker detection methods, and not as a way to assess model performance. Examples of the resulting images are depicted in Fig.~\ref{fig:flyingarucov2}.

To train our detector, we have split the dataset into two subsets, one for train and the other for validation. The training set contains $2000$ images, and the validation set contains $500$ images. The markers included in each subset are cropped to obtain the corner regression and marker decoding datasets: the markers from the training subset will be used as the training dataset, and the markers from the validation subset will be used as the validation dataset.

\subsection{Shadow-ArUco dataset} \label{ssec:shadowaruco}

To test the methods we have developed and compare them against other methods, we have obtained a second dataset in real-world conditions, dubbed the Shadow-ArUco dataset. This dataset was initially presented in IbPRIA'23~\cite{berral2023ibpria}.

For the construction of this dataset, we have attached multiple markers to the walls in the corner of a darkened room (we closed the blinds and turned off the lights); then, a video containing moving shadows and lighting patterns is projected over the corner. The entire scene is then recorded from different points of view, obtaining multiple video sequences. 

To estimate the ground-truth labels of the markers in the scene, we have used a semi-manual method: first, the frames are processed using the traditional ArUco method, setting the error correction bits (ECB) parameter equal to zero to get only the detections with the highest confidence, as this weeds out most false detections. Since the camera is fixed during each video sequence, marker positions are consistent across frames. If a marker is not detected in a particular frame, its corners and ID can be retrieved from the nearest frame where it was detected. As a last resort, if a marker is not detected, we manually add the position of its corners and ID and then copy this information for every frame in the sequence.
%However, as the camera is fixed while recording each video sequence, marker positions are shared for every frame of the sequence; if a marker is not detected in a frame, we can get its corners and ID from the closest frame in which the marker has been detected. 

\begin{figure}[t]
\includegraphics[width=1.\linewidth]{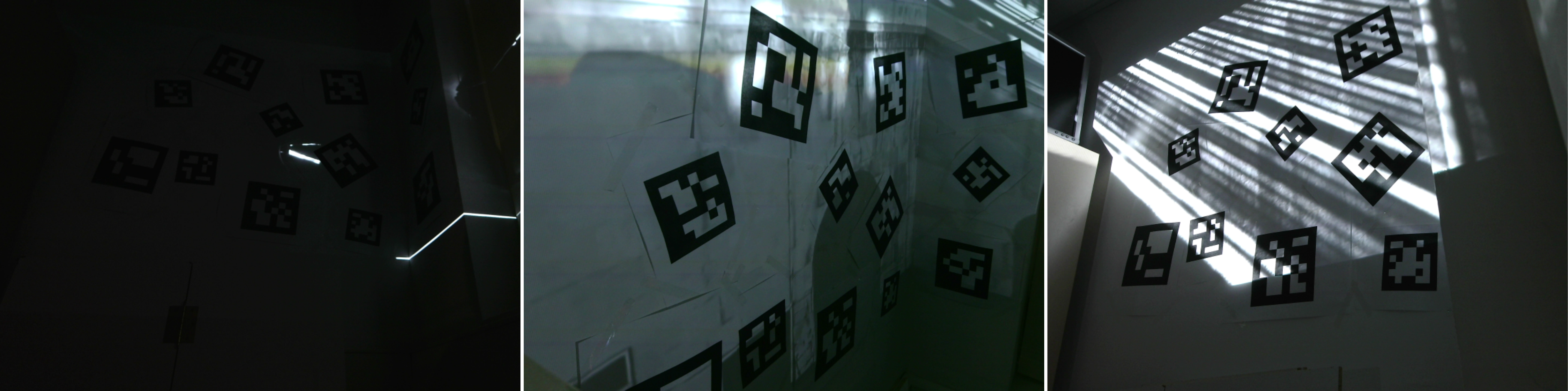}
\vspace{0.5mm}
\caption{\textbf{Shadow-ArUco dataset.} A video containing complex lighting patterns is projected on the corner of a room, over whose walls multiple patterns have been attached; the complex lighting makes marker detection difficult for classical methods. The scene is recorded from multiple fixed camera positions. (Best viewed in digital format)}
\label{fig:shadowaruco}
\end{figure}

We have recorded the scene from six different points of view, obtaining six different video sequences for a total of $8652$ frames, with a resolution of $800\times 600$. Examples of the images contained in this dataset are depicted in Fig.~\ref{fig:shadowaruco}.

\subsection{DeepTag ArUco dataset} \label{ssec:deeptag_aruco}

\begin{figure}[t]
\includegraphics[width=1.\linewidth]{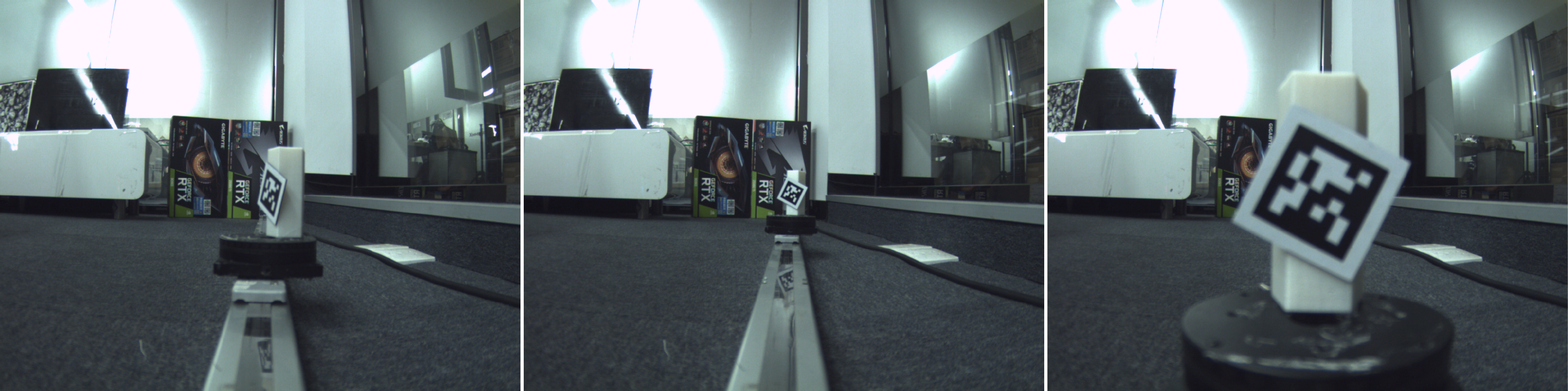}
\vspace{0.5mm}
\caption{\textbf{DeepTag ArUco dataset.} A single ArUco marker is photographed in a fixed environment at varying distances and orientations w.r.t. the camera. (Best viewed in digital format)}
\label{fig:deeptag}
\end{figure}

Finally, to provide a fair comparison against the model proposed in \cite{zhang2023deeptag}, we have tested our methods on the DeepTag dataset built by the authors of the method. The dataset can be obtained at the DeepTag project page \footnote{DeepTag project page: \url{https://herohuyongtao.github.io/research/publications/deep-tag/}}. This dataset comprises photographs of different fiducial markers in fixed positions at varying distances and orientations with respect to the camera. For each position, a total of $100$ photos are obtained. While the authors do this to take into account sensor jitter, images obtained for a given marker type from a particular position do not have significant differences; nevertheless, we will consider each independent frame as a different image for testing purposes. Some examples from this dataset appear in Fig.~\ref{fig:deeptag}.

As the original dataset contains examples for multiple marker families, we restrict our comparison only to those images containing ArUco markers. Taking this into account, the resulting dataset contains markers photographed with view angles of $\{0, 10, 20, 30, 40, 50, 60, 65, 70,  75\}$ degrees at distances of $\{10, 20, 30, 40, 50, 60, 70, 80, 90,  100\}$ cm., for a total of $100$ different positions with respect to the camera,  for a total of $10000$ frames ($100$ frames per position), with a resolution of $1280 \times 960$. We will restrict our comparison to the \texttt{general} dataset (i.e., we will not measure the performance of our methods in the images containing Gaussian noise or simulated motion blur, as these tasks fall out of the scope of this work). Please note that each image contains a single marker and is the same for every image (i.e., the encoded ID is the same).
%\section{Proposed Methods} 
\section{Proposed DeepArUco++ Framework} \label{sec:methods}

Our approach to marker detection and identification is built around a pipeline obtained by dividing the problem into simpler tasks. The initial task involves detecting the markers and their bounding boxes (see Section~\ref{ssec:detector}). This is done without finding the ID of the marker; i.e., the overall position of the marker is found without decoding it. Then, for every detected marker, the precise location of its corners is determined (see Section~\ref{ssec:corners}): this allows us to determine the exact location of the marker to rectify it, making decoding it easier. Finally, the markers are decoded by determining the value of each bit and comparing the extracted bits against the ArUco codes dictionary (see Section~\ref{ssec:decoder}).

\subsection{Bounding-box level detection} \label{ssec:detector}

The first step in our pipeline is the broad-brush location of the markers contained in the frame. This is done using a model based on the YOLOv8 object detector; we have considered both its nano (\texttt{yolov8n}) and small (\texttt{yolov8s}) sized variants, starting from the provided pre-trained models. These models were trained using the Flying-ArUco v2 dataset, presented in Section~\ref{ssec:flyingaruco}.

While we have started from off-the-shelf YOLOv8 models, improvements regarding robustness towards challenging lighting conditions stem from the use of the dataset designed in Section \ref{ssec:flyingaruco} as training data.
As a form of offline data augmentation, we have considered multiple approaches. The most general, applied for every augmented frame, consists of multiplying each image by a synthetic gradient with an arbitrary angle, with random minimum and maximum values ranging from $0.0$ to $2.0$; this is done to simulate an even broader range of lighting conditions. Additionally, other transformations have been considered, such as shifting the color of every frame by a small amount (i.e., applying a particular ``tint'' to the input image), applying full-frame Gaussian blur (with an overall probability of $20\%$) or adding a varying amount of Gaussian noise to the frame. In total, nine additional augmented samples are created for each non-augmented sample (the size of the dataset grows by a factor of $10\times$). Different transformations and model size combinations have been used to find the best-performing configuration when testing on both Shadow-ArUco and DeepTag ArUco datasets. No online data augmentation has been used, and other parameters, such as the loss function, have been kept as default.

\subsection{Corner regression} \label{ssec:corners}

The second step in our pipeline is the precise location of the corners of each marker to better fit the actual shape and dimensions of the detected marker. Two alternative approaches have been considered for this task: (a) direct corner regression (i.e. directly regressing the coordinates of each corner) and (b) heatmap-based corner regression (i.e. directly locating the corners in the input image, to later obtain the precise coordinates). Please note that as presented in this work the two methods are mutually exclusive: only one of them can be used at a time, as they provide the same functionality in our proposed pipeline.

\subsubsection{Direct corner regression} \label{sssec:direct_regression}

\begin{figure}[t]
    \centering
    \includegraphics[width=\linewidth]{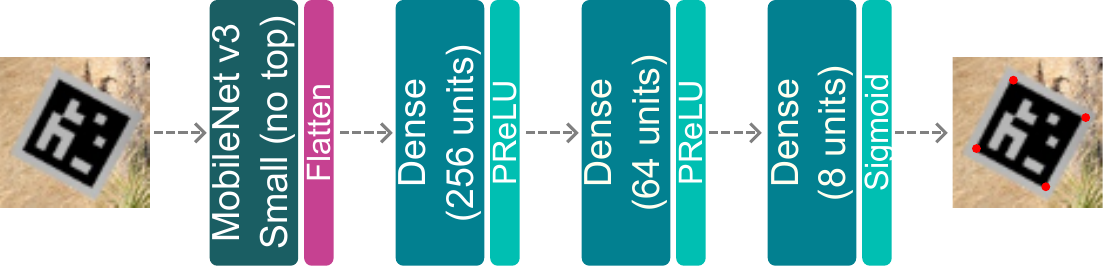}
    \vspace{0.5mm}
    \caption{\textbf{Corner regressor architecture.} The model's input is a $64\times64$ color image containing a marker; the output is a vector containing eight values corresponding to the $x$ and $y$ coordinates of the four corners of the marker in counterclockwise order. Best viewed in digital format.}
    \label{fig:regressor_direct}
\end{figure}

The first approach we have considered for the task of corner regression is that of directly regressing the coordinates of each corner, i.e., predicting the eight different values corresponding to the $(x,y)$ coordinates of each corner of a given marker (two values for each corner). We have used a model based around a pre-trained MobileNetV3 backbone (in its \texttt{MobileNetV3Small} variant). We removed the original, fully-connected top from this backbone and kept the rest as a feature extractor. Then, we added a custom top composed of two fully-connected layers with sizes $256$ and $64$, with a PReLU activation function, and a third fully-connected layer, with size $8$ (two coordinates for each corner) and a sigmoid activation function as the output. The network expects a $64 \times 64$ pixel color image as input and outputs four pairs of coordinates within the $[0, 1]$ range, representing the positions of the corners relative to the width/height of the input image. A summary of the architecture appears in Fig.~\ref{fig:regressor_direct}.
%As input, the network expects a $64 \times 64$ pixels color image, and the model will output four pairs of coordinates with values in the $[0, 1]$ range (i.e., the position of the corners is represented using values relative to the width/height of the input image). 

\begin{figure}[t]
    \centering
    \includegraphics[width=\linewidth]{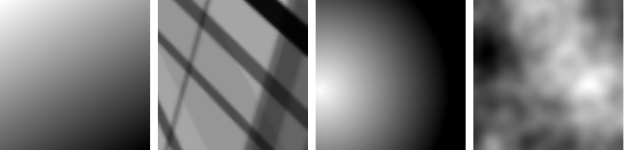}
    \vspace{0.5mm}
    \caption{\textbf{Lighting patterns used for data augmentation.} From left to right: simple gradients, lines, circular patterns, and Perlin noise. Multiple patterns can be combined (i.e., multiplied) to achieve even more complex augmentations.}
    \label{fig:lighting_patterns}
\end{figure}

To train the models, the markers from the Flying-ArUco v2 dataset (see Section~\ref{ssec:flyingaruco}) are cropped by using their ground-truth bounding boxes (previously expanded by a margin of 20\% over the original width at both left and right and a margin of $20\%$ of the original height at both top and bottom), and then resizing to a size of $64 \times 64$. Markers from the training subset (after offline data augmentation) will be used to train the model, and markers from the validation subset will be used to verify the progress of the training. As a form of online data augmentation, images will be rotated by a multiple of $90^\circ$, mirrored horizontally or vertically, and multiplied by a lighting pattern composed of lines, Perlin noise, and/or circular shadows (see Fig.~\ref{fig:lighting_patterns}); multiple transformations can be applied to the same sample. Labels associated with each input must also be transformed for consistency with the transformations applied to the crops (i.e., the coordinates of each corner and their order must reflect the rotations and mirroring applied to the markers). As the loss function, we will use the mean average error (MAE) between the predicted coordinates and the (properly transformed) ground-truth corners' positions.

In inference time, the input to the model will be obtained by expanding the bounding boxes found by the marker detector as we did to obtain the training crops. Then, markers are cropped and resized to a size of $64 \times 64$, and the resulting image is used as the input to the corner regressor. Final positions are computed by multiplying each output by the width/height of the bounding box as corresponding, then offsetting each position by the minimum $x$ and $y$ values of the expanded bounding box.

\subsubsection{Heatmap-based corner regression} \label{sssec:heatmap_regression}

\begin{figure}[t]
    \centering
    \begin{subfigure}[t]{0.387\textwidth}
        \centering
        \includegraphics[width=\textwidth]{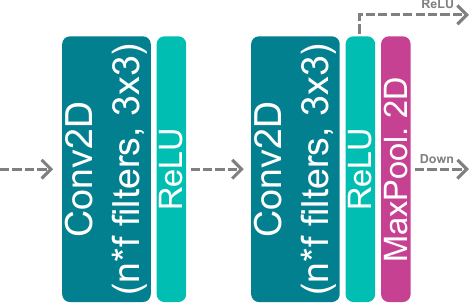}
        \caption{Downsampling block.}
    \end{subfigure}
    \hfill
    \begin{subfigure}[t]{0.513\textwidth}
        \centering
        \includegraphics[width=\textwidth]{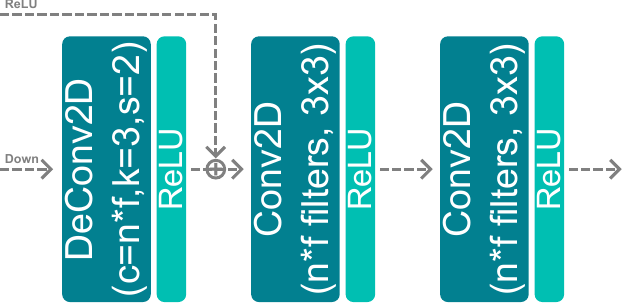}
        \caption{Upsampling block.}
    \end{subfigure}
    \par\bigskip
    \begin{subfigure}[t]{\textwidth}
        \centering
        \includegraphics[width=\textwidth]{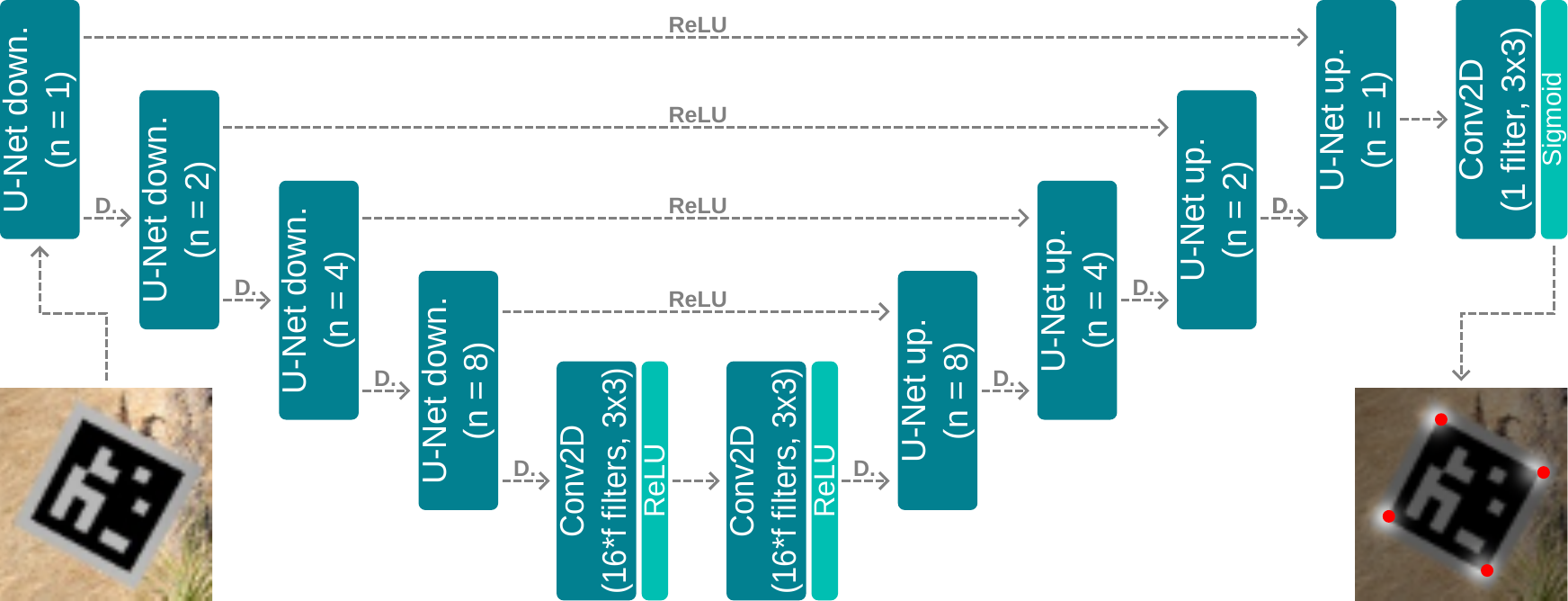}
        \caption{Full architecture.}
    \end{subfigure}
    \vspace{0.5mm}
    \caption{\textbf{Heatmap-based corner regressor architecture.} The model's input is a $64\times64$ color image containing a marker; the output is a heatmap containing \textit{blobs} in the areas corresponding to the detected corners. In strided convolutions (\texttt{DeConv2D}), $c$ stands for the number of filters, $k$ stands for the kernel size ($k \times k$), and $s$ stands for the used stride size ($s \times s$). In (b), $+$ signals a concatenation of inputs. In (c), $D.$ stands for \textit{Down} (as in (a) and (b), abbreviated for readability). Best viewed in digital format.}
    \label{fig:regressor_hmap}
\end{figure}

As a second approach to precise corner location, we have implemented a heatmap-based approach. Taking inspiration from keypoint-based methods, such as those used in human pose estimation and facial landmarks location \cite{alphapose2016, mediapipe2019}, we compute for each coordinate a Gaussian heatmap, reflecting a probability density function, following a normal distribution whose mean corresponds to the position of the encoded corner. Then, the maps corresponding to each corner are combined to obtain a single map reflecting the most probable locations of every corner in the image. As the architecture for this model, we have used a U-Net~\cite{unet_paper}, as similar approaches have been used with relative success in keypoint detection tasks \cite{unet1,unet2}; marker corners can be considered as a type of keypoint. A summary of the architecture used in this approach appears in Fig.~\ref{fig:regressor_hmap}.

As a heatmap built in such a way is mainly composed of values near zero, two corrections must be made to allow our models to learn from these images. The first is to try and maximize the number of non-zero values in the heatmap: to do so, we have increased the values in the covariance matrix by a factor of ten. Additionally, we have used a weighted loss function based on the mean squared error (MSE), weighted in such a way that the error near the corners we aim to predict is valued up to ten times as much as the error far from the corners. See Equations~\ref{eq:w_mse_1} and~\ref{eq:w_mse_2} for a better understanding of the loss calculation, where $Y$ stands for the target heatmap to predict, $r$ and $c$ stands for the number of rows and columns in the output, respectively, $i$ stands for the row number (from $0$ to $r - 1$), $j$ stands for the column number (from $0$ to $c - 1$) and $\mathcal{L}_h$ stands for the weighted loss function. 

\begin{equation} \label{eq:w_mse_1}
weights = \frac{Y - Y_{min}}{Y_{max} - Y_{min}} \cdot 9 + 1
\end{equation}

\begin{equation} \label{eq:w_mse_2}
\mathcal{L}_h(Y,\hat{Y}) = \frac{1}{r \cdot c}\sum_{i}^{r}\sum_{j}^{c}(\hat{Y}_{i,j} - Y_{i,j})^{2} \cdot weights_{i,j}
\end{equation}

To train the models, we have used the markers from the Flying-ArUco v2 dataset (see Section~\ref{ssec:flyingaruco}) cropped and augmented as described in Section \ref{sssec:direct_regression}, but using the heatmaps computed from the ground-truth corners as the labels to predict. These markers are also normalized to the $[0, 1]$ range.

On inference time, the input for the models will be obtained the same way as for the direct regression approach. However, to use the outputs, we have to compute the corner positions from the regressed heatmaps. To do so, we propose the following method: first, we extract the \textit{blobs} (areas with high values relative to the rest of the map) from the regressed map using the blob extractor provided by the OpenCV library, tuned to the expected blob area (i.e., finding blobs with an area similar to that of the blobs in the training heatmaps). The average value is computed for every detected blob and used as its score. The top four blobs are used to calculate the refined corner positions (please note that it is possible for the method to locate less than four corners). From every blob, a mask is computed (to discard every value outside the detected blob). The final position ($x$, $y$) is computed as the sum of the coordinates of each non-zero valued position, weighted by the contribution of the value in that position to the sum of all values in the blob. For a better understanding of the position calculation from a given blob, see Eq.~\ref{eq:xy_from_blob}), where $blob$ stands for the heatmap after zero masking every pixel not belonging to the blob of interest and $i$ and $j$ indicate a row and a column in the heatmap, respectively.

\begin{equation} \label{eq:xy_from_blob}
x,~y = \sum_{i,j}j\cdot\frac{blob_{i,j}}{\sum_{i,j}{blob_{i,j}}},~\sum_{i,j}i\cdot\frac{blob_{i,j}}{\sum_{i,j}{blob_{i,j}}}
\end{equation}

\subsection{Marker decoding} \label{ssec:decoder}

\begin{figure}[t]
    \centering
    \includegraphics[width=\linewidth]{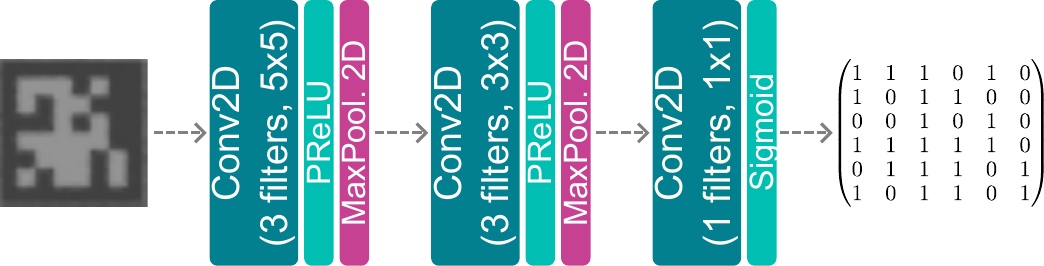}
    \vspace{0.5mm}
    \caption{\textbf{Marker decoder architecture.} The model's input is a $32\times32$ grayscale image containing a rectified marker; the output is a matrix of size $6\times6$, containing the prediction for each bit in the input picture, in the range [0-1].}
    \label{fig:decoder}
\end{figure}

The last step in our pipeline is extracting the bits encoded in each marker. We will use a simple architecture composed of three convolutional layers (see Fig.~\ref{fig:decoder}). As input, the model expects a $32\times32$ pixels grayscale image, normalized with respect to its maximum and minimum values to the $[0, 1]$ range. As its output, it returns a $6\times6$ array of values in the range $[0, 1]$, which, after rounding, can be interpreted as the decoded bits. 

We have used the markers from the Flying-ArUco v2 dataset to train the model, cropped from their ground-truth corner positions. As a form of online data augmentation, we shift ground-truth coordinates by up to $5\%$ in each direction before cropping the markers; this is done to simulate ``imperfect'' corner detections to obtain a more robust model; additionally, we have used the same lighting, flipping, and rotation transformations as for the training of the corner refinement models (see Section \ref{ssec:corners}). We used the MAE between the predicted values and the ground-truth encoded bits as a loss function.

On inference time, the input for the model is obtained using the corners predicted by the corner refinement model in the previous step; then, after locating the four corners, the marker is rectified through a homography, by finding the transformation which maps each corner to its corresponding corner in a completely flat square. After rectifying, the values in the resulting image are normalized to the $[0, 1]$ range and used as input for the decoder network. Then, the output of the decoder is flattened and rounded to the nearest integer (i.e., $0$ or $1$), and the resulting sequence of bits is compared against every possible code in the ArUco dictionary. As there are four possible orientations for the marker, the comparison is performed four times, rotating $90^\circ$ each time (i.e., the output $6\times6$ matrix is rotated, flattened, and finally compared). The marker is then assigned the ID corresponding to the ArUco code with the smallest Hamming distance with respect to the encoded bits in any orientation. Finally, by defining a maximum Hamming distance as a threshold, we can discard false detections based on the extracted ID; however, this will likely translate into a smaller recall (i.e., some correctly detected markers may be discarded).
\section{Experiments and Results} \label{sec:results}

\subsection{Metrics and methodology}

To validate our methods, we have used a combination of multiple metrics, testing each stage of the pipeline individually. To measure the capabilities of the different marker detectors obtained in this work, we have used a Precision-Recall curve, starting from the detections and ground-truth marker locations in the form of bounding boxes. We consider a detection correct if it overlaps with some ground-truth bounding box with an \textit{Intersection-over-Union} (IoU) score of at least $0.5$. The models will be tested over the Shadow-ArUco dataset described in Section~\ref{ssec:shadowaruco}.

To test the performance of our corner refinement models, we will measure the average per-corner error, computed as the distance in pixels between a predicted corner and its nearest ground-truth corner, and the time spent per frame, measured in milliseconds. Additionally, for the methods based on heatmaps, we will get the percentage of markers for which one, two, three, or four corners have been found (as the described method may find less than four corners). As input for the models, we will use the ground-truth marker annotations from the Shadow-ArUco dataset, cropped and expanded as described in Section~\ref{sssec:direct_regression}.

Marker decoding was assessed using a different Precision-Recall curve, each point representing the precision and recall values obtained after filtering the output of the decoder using a different threshold. For this test, recall is the ratio between the number of detected markers remaining after discarding those for which the closest ArUco code is found at a distance greater than the target threshold and the total number of ground-truth markers. Precision will be computed as the ratio between the number of correctly identified markers at a distance below that target threshold and the number of markers remaining after the previous filtering.

Finally, we will test the capabilities of the entire pipeline at different configurations and compare them against previous state-of-the-art fiducial marker detection methods (classic ArUco and DeepTag). We will also consider the previous iteration of this work in~\cite{berral2023ibpria} as a baseline. In this comparison, we will use Precision-Recall curves to test the detection capabilities of each method. DeepArUco++ and baseline DeepArUco will have their detections filtered by thresholding at the decoding step. Additionally, a numerical, quantitative comparison will be provided, measuring the area under the Precision-Recall curve (AUC-PR) for every method, the minimum precision and maximum recall values (useful to test the detection capabilities of the different methods under no additional filtering), the average per-corner error and the percentage of accuracy, computed as the number of correctly identified markers with respect to the number of true detections (i.e., detected markers which correspond with a real, ground truth marker) after filtering. A comparison of the time spent per frame with each method will be provided, with a per-step breakdown for the different DeepArUco++ configurations.

\subsection{Implementation details}

We have used Ultralytics YOLOv8~\cite{yolov8} to train our marker detector. Both our corner regressor and marker decoder have been implemented using TensorFlow. Our experiments were run on a computer with an Intel(R) Core(TM) i7-11700F CPU and an NVIDIA GeForce RTX 3090(R) GPU. 

We trained the marker detector using rectangular training (see Ultralytics YOLO documentation\footnote{YOLOv8 documentation: \url{https://docs.ultralytics.com/usage/cfg/\#train-settings}}), with an IoU threshold of 0.5 for non-maximum suppression, automatic batch size computation \new{(it varies between executions, based on the available amount of GPU memory)}, and a maximum of 1000 epochs, with early stopping after 10 epochs without validation improvement. \new{The rest of the parameters used to control the training of the detector} were set to default; \new{a learning rate of $1e-2$ and automatic selection of optimizer (using SGD in our runs, with a momentum of 0.9).} For the corner refinement models, we used the Adam optimizer with a batch size of 32, a maximum of 1000 epochs, early stopping after 10 epochs without improvement, and learning rate halving after 5 epochs without improvement (initial learning rate: $1e-3$). Refer to Section \ref{ssec:corners} for loss functions. The marker decoder was also trained with the Adam optimizer, batch size of 32, and up to 1000 epochs, with early stopping after 20 epochs and learning rate halving after 10 epochs without improvement (initial learning rate: $1e-3$). The loss function used was the mean average error (MAE). We refer the reader to the public source code of DeepArUco++ (see Abstract) for further implementation details.

\subsection{Method development} \label{ssec:quantresults}

% Marker detector
\subsubsection{Marker detection} \label{sssec:results_detection}

\begin{figure}[t]
    \centering
    \begin{subfigure}[t]{.49\textwidth}
        \centering
        \includegraphics[width=\textwidth]{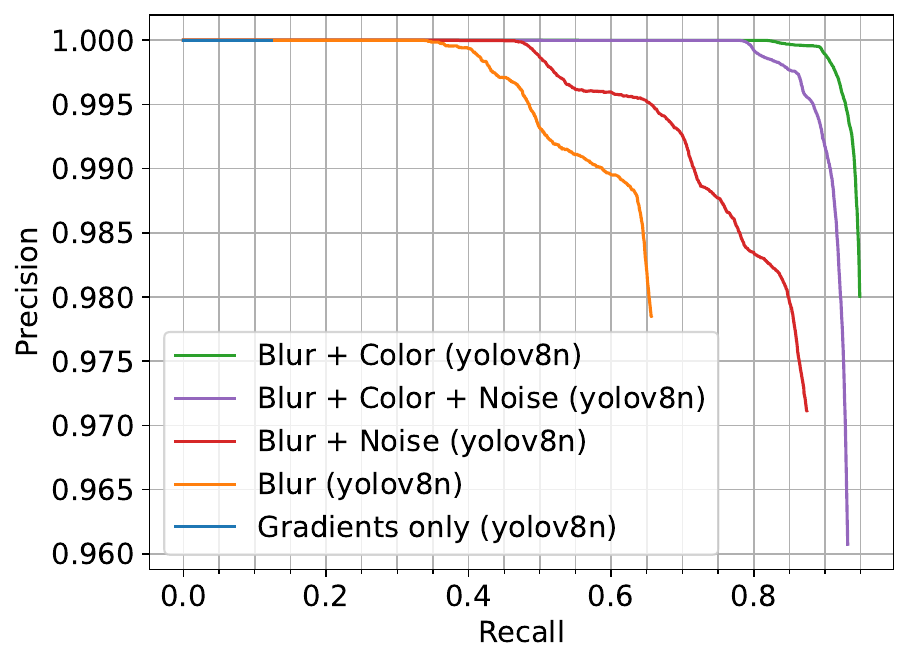}
        \caption{}
    \end{subfigure}
    \begin{subfigure}[t]{.49\textwidth}
        \centering
        \includegraphics[width=\textwidth]{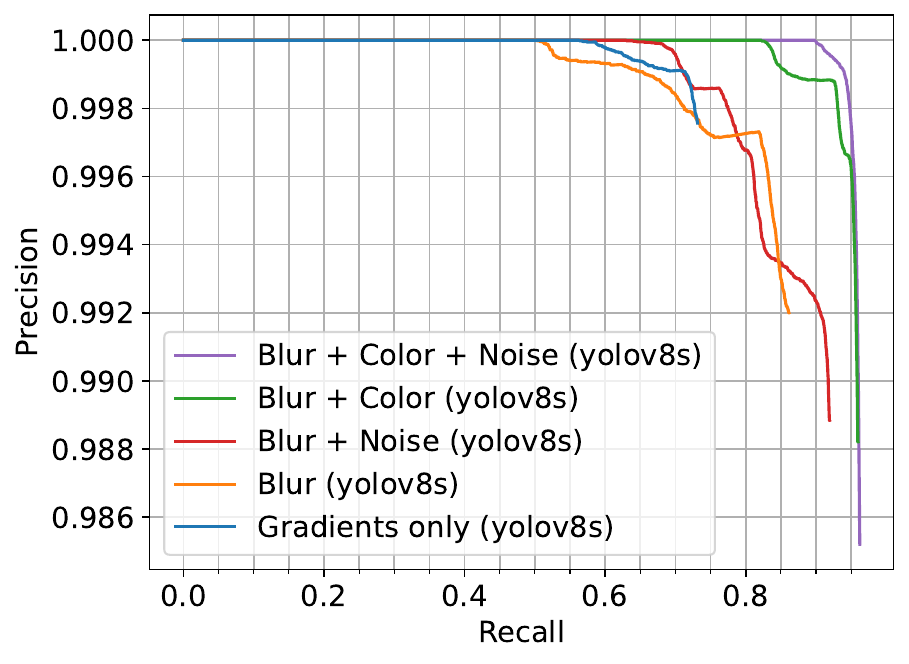}
        \caption{}
    \end{subfigure}
    \caption{\textbf{Comparison of different training configurations on Shadow-ArUco dataset.} \textbf{(a)} Precision-Recall curve for models based on \texttt{yolov8n} (nano variant), \textbf{(b)} Precision-Recall curve for models based on \texttt{yolov8s} (small variant). Legend reflects the options used for data augmentation in each training; \textit{Color} stands for color shift; \textit{Noise} stands for Gaussian noise. Best viewed in digital format.}
    \label{fig:detector_comparison}
\end{figure}

From the methodology described in Section \ref{ssec:detector}, we have trained $10$ different marker detectors using different combinations of augmentation options and base model sizes. In Fig.~\ref{fig:detector_comparison}, we compare the Precision-Recall (PR) curves obtained when testing over the Shadow-ArUco dataset from Section~\ref{ssec:shadowaruco}.

We can extract the following conclusions from the comparison in Fig.~\ref{fig:detector_comparison}. From a Precision-Recall standpoint, the optimal approach appears to be training a model from a pre-trained \texttt{yolov8s} model, using blur, color shift and Gaussian noise to produce augmented samples (see Section~\ref{ssec:detector}), with an AUC of $0.9622$. 
However, if we consider the method's efficiency, using a model based on \texttt{yolov8n} will likely result in a better throughput. Training from a pre-trained \texttt{yolov8n}, and using both blur and color shift as augmentation options, the resulting model has an AUC of $0.9485$, not far from the best \texttt{yolov8s}-based model. 
Overall, it seems that the best combination of options when considering both model sizes is to use just blur and color shift, as it yields the best performance when training from \texttt{yolov8n}, and gets the second place when training from \texttt{yolov8s} (with 0.9592, versus 0.9622 when using every option).
%maximum recall to the best settings found for the \texttt{yolov8s} base (0.9488 vs 0.9594), while keeping a reasonable minimum precision (0.9801 vs 0.9882).

% Corner regression
\subsubsection{Corner regression} \label{sssec:results_regression}

\begin{figure}[t]
    \centering
    \hfill
    \begin{subfigure}[t]{.47\textwidth}
        \centering
        \includegraphics[width=\textwidth]{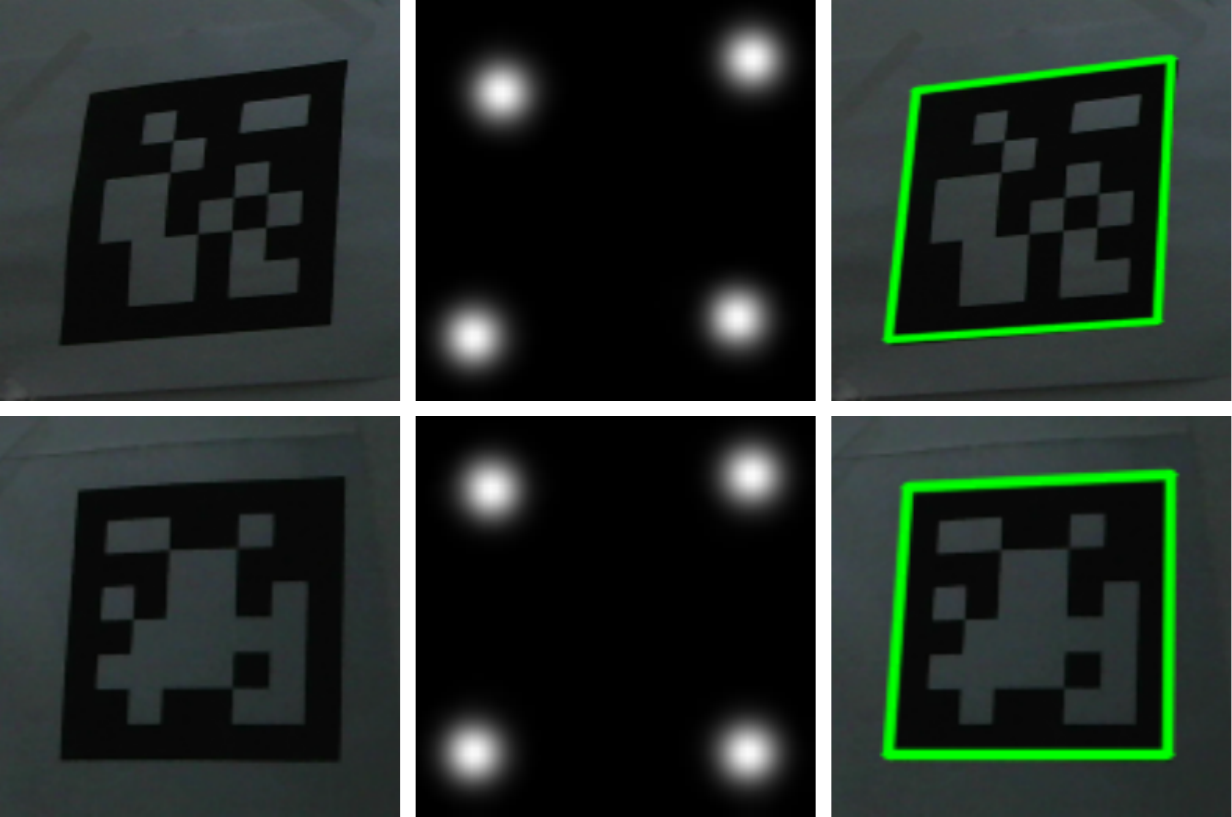}
    \end{subfigure}
    \hfill
    \begin{subfigure}[t]{.47\textwidth}
        \centering
        \includegraphics[width=\textwidth]{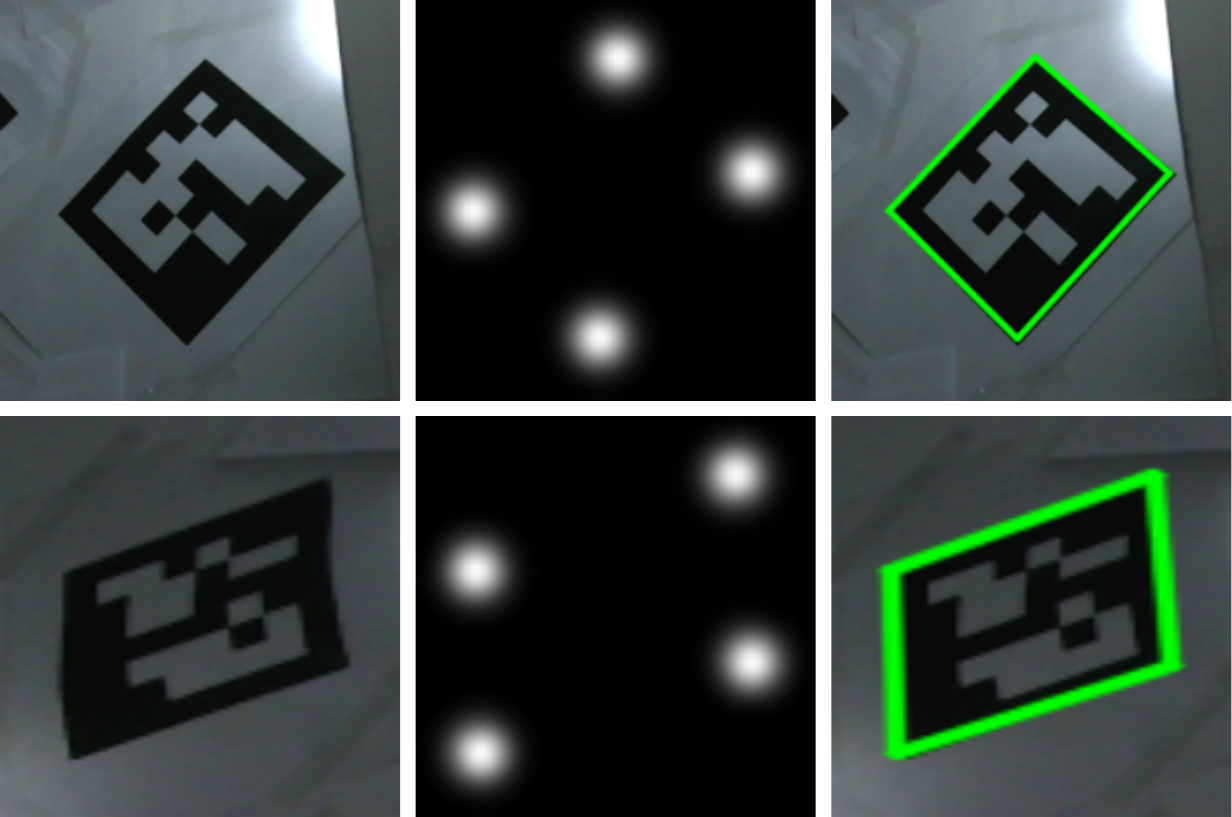}
    \end{subfigure}
    \hfill
    \caption{\textbf{Examples of heatmaps and marker locations obtained using our method.} For each block, the left row shows the original crops, the middle row shows the obtained heatmaps and the right row shows the obtained precise marker location. Best viewed in digital format.}
    \label{fig:hmap_examples}
\end{figure}

Following the methodology described in Section~\ref{ssec:corners}, we have trained five different corner refinement models; one based on the MobileNetV3-based architecture from Section~\ref{sssec:direct_regression}, which performs the direct regression of the coordinates of the four corners, and the rest based on the U-Net based architecture from Section~\ref{sssec:heatmap_regression}, which locates the corners in the form of a heatmap (five different configurations have been tested, based on the number of filters on the first convolutional layer). The models have been trained on the cropped markers obtained from the Flying-ArUco v2 dataset, cropping after adding blur and color shift to the whole image, as it yielded the best overall results for the training of the detectors on Section~\ref{sssec:results_detection} (online data augmentation has also been applied on a crop-by-crop basis, following the method on Section~\ref{sssec:direct_regression}). Some examples of heatmaps generated using our method are shown in Fig.~\ref{fig:hmap_examples}.

%\begin{figure}[t]
%    \centering
%    \includegraphics[width=\textwidth]{figures/heatmap_examples.pdf}
%    \caption{\new{\textbf{Examples of heatmaps and marker locations obtained using our method.} The left row shows the original crops, the middle row shows the obtained heatmaps and the right row shows the obtained precise marker location. Best viewed in digital format.}}
%    \label{fig:hmap_examples}
%\end{figure}

Please note that in our design for the heatmap-based refinement model, it is possible to locate less than four corners; also, for both the heatmap-based approach and the direct regression approach, we discard detections with their nearest ground-truth corner found at a distance greater than 5 pixels. 
A comparison of the results obtained by the different models, testing over the crops obtained from the Shadow-ArUco dataset (using the expanded ground-truth bounding boxes), appears in Table~\ref{tab:regressor_comparison}.

\begin{table}[t]
    \centering
    \caption{\textbf{Comparison of different corner refinement models on Shadow-ArUco dataset.} \textit{Error} stands for the mean Euclidean distance between a predicted corner and its nearest ground-truth corner (in pixels); \textit{Time} refers to the mean inference time per frame, in milliseconds (with $10.90 \pm 1.33$ markers per frame); $f$ refers to the number of filters in the first convolutional layer (when applicable). Percentages of markers for which a certain number of corners have been found have been computed over the total number of markers in the Shadow-ArUco dataset. Note that \textit{Error} only takes into account detected corners (could be less than 4 corners). For metrics noted with $\downarrow$, lower is better.}
    \small
    \begin{tabular}{|c|c|c|c|c|c|c|c|}
        \cline{2-6}
        \multicolumn{1}{c|}{} & \multicolumn{5}{c|}{\# of found corners} & \multicolumn{2}{c}{} \\
        \hline
        \multirow{2}{*}{\makecell[c]{\textbf{Method}}} & \multirow{2}{*}{\makecell[c]{\textbf{0}}} & \multirow{2}{*}{\makecell[c]{\textbf{1}}} & \multirow{2}{*}{\makecell[c]{\textbf{2}}} & \multirow{2}{*}{\makecell[c]{\textbf{3}}} & \multirow{2}{*}{\makecell[c]{\textbf{4}}} & \multirow{2}{*}{\makecell[c]{\textbf{Error $\downarrow$}}} & \multirow{2}{*}{\makecell[c]{\textbf{Time $\downarrow$}}}\\
        & & & & & & &\\
        \hline
        \hline
        \multirow{3}{*}{\makecell[c]{Direct}} & \multirow{3}{*}{\makecell[c]{0.74\%}} & \multirow{3}{*}{\makecell[c]{2.43\%}} & \multirow{3}{*}{\makecell[c]{5.70\%}} & \multirow{3}{*}{\makecell[c]{10.97\%}} & \multirow{3}{*}{\makecell[c]{80.16\%}} & \multirow{3}{*}{\makecell[c]{1.55 $\pm$\\ 1.01}} & \multirow{3}{*}{\makecell[c]{\textbf{3.29 $\pm$}\\ \textbf{3.69}}}\\
        & & & & & & &\\
        & & & & & & &\\
        \hline
        \hline
        \multirow{3}{*}{\makecell[c]{Heatmap\\($f = 2$)}} & \multirow{3}{*}{\makecell[c]{0.01\%}} & \multirow{3}{*}{\makecell[c]{0.23\%}} & \multirow{3}{*}{\makecell[c]{2.33\%}} & \multirow{3}{*}{\makecell[c]{9.50\%}} & \multirow{3}{*}{\makecell[c]{87.93\%}} & \multirow{3}{*}{\makecell[c]{1.61 $\pm$\\ 0.85}} & \multirow{3}{*}{\makecell[c]{6.24 $\pm$\\ 3.45}}\\
        & & & & & & &\\
        & & & & & & &\\
        \hline
        \multirow{3}{*}{\makecell[c]{Heatmap\\($f = 4$)}} & \multirow{3}{*}{\makecell[c]{0.00\%}} & \multirow{3}{*}{\makecell[c]{0.07\%}} & \multirow{3}{*}{\makecell[c]{0.79\%}} & \multirow{3}{*}{\makecell[c]{9.65\%}} & \multirow{3}{*}{\makecell[c]{89.48\%}} & \multirow{3}{*}{\makecell[c]{1.51 $\pm$\\ 0.77}} & \multirow{3}{*}{\makecell[c]{6.34 $\pm$\\ 3.59}}\\
        & & & & & & &\\
        & & & & & & &\\
        \hline
        \multirow{3}{*}{\makecell[c]{Heatmap\\($f = 8$)}} & \multirow{3}{*}{\makecell[c]{0.00\%}} & \multirow{3}{*}{\makecell[c]{0.06\%}} & \multirow{3}{*}{\makecell[c]{0.89\%}} & \multirow{3}{*}{\makecell[c]{8.63\%}} & \multirow{3}{*}{\makecell[c]{90.41\%}} & \multirow{3}{*}{\makecell[c]{1.51 $\pm$\\ 0.72}} & \multirow{3}{*}{\makecell[c]{6.44 $\pm$\\ 3.82}}\\
        & & & & & & &\\
        & & & & & & &\\
        \hline
        \multirow{3}{*}{\makecell[c]{Heatmap\\($f = 16$)}} & \multirow{3}{*}{\makecell[c]{0.00\%}} & \multirow{3}{*}{\makecell[c]{0.06\%}} & \multirow{3}{*}{\makecell[c]{0.49\%}} & \multirow{3}{*}{\makecell[c]{8.68\%}} & \multirow{3}{*}{\makecell[c]{90.77\%}} & \multirow{3}{*}{\makecell[c]{1.51 $\pm$\\ 0.71}} & \multirow{3}{*}{\makecell[c]{6.57 $\pm$\\ 4.50}}\\
        & & & & & & &\\
        & & & & & & &\\
        \hline
        \multirow{3}{*}{\makecell[c]{Heatmap\\($f = 32$)}} & \multirow{3}{*}{\makecell[c]{0.00\%}} & \multirow{3}{*}{\makecell[c]{0.04\%}} & \multirow{3}{*}{\makecell[c]{0.52\%}} & \multirow{3}{*}{\makecell[c]{8.44\%}} & \multirow{3}{*}{\makecell[c]{\textbf{91.00\%}}} & \multirow{3}{*}{\makecell[c]{\textbf{1.44 $\pm$}\\ \textbf{0.69}}} & \multirow{3}{*}{\makecell[c]{6.87 $\pm$\\ 4.63}}\\
        & & & & & & &\\
        & & & & & & &\\
        \hline
    \end{tabular}
    \label{tab:regressor_comparison}
\end{table}

From the results in Table~\ref{tab:regressor_comparison}, we observe that the heatmap-based approach clearly improves the corner detection rate. Additionally, it improves the average per-corner error (mostly reducing the variability), especially with $f = 32$. On the other hand, it has a much higher inference time than the direct regression approach. This is caused by the need to extract the actual corner positions from the heatmap; much optimization is required on the implementation side. Considering this, we recommend using the heatmap-based approach when maximum performance is required, leaving the direct-regression approach for scenarios where throughput is a concern.

%However, we can see how the method based on heatmaps has a much higher inference time than the direct regression method. This is caused by the need to extract the actual corner positions from the heatmap (e.g., the mean time required for inference only for the model with $f = 8$ is $3.11 \pm 5.40$ milliseconds); much optimization is required on the side of implementation. On the other hand, if we focus only on the number of ``perfect'' marker locations (defined here as markers with an IoU with respect to their nearest ground-truth marker of 0.9 or higher), using the heatmap-based approach leads to a greater yield, especially with $f \geq 4$; from the perspective of the average error (i.e., computed only over the detected corners, while excluding the missing ones), the heatmap-based approach can also lead to some improvement when compared to the direct regression, specifically the configurations with $f = 8$ or $f = 16$. However, we must keep in mind that there is a chance of the model not finding every corner of a given marker, even though it is small, at least on our test dataset. 

% Marker decoding
\subsubsection{Marker decoding} \label{sssec:results_decoding}

\begin{figure}[t]
    \centering
    \includegraphics[width=0.75\textwidth]{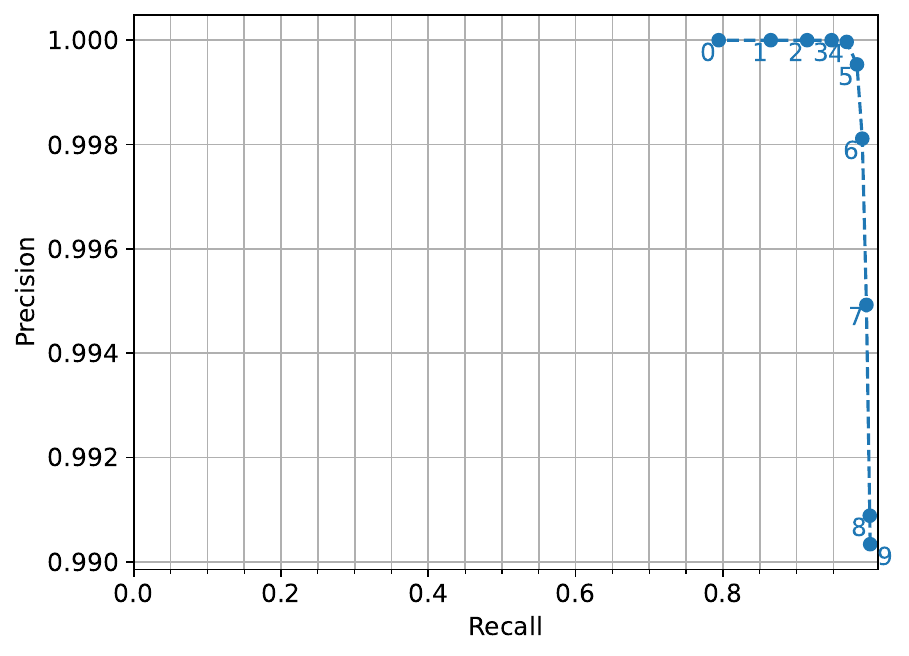}
    \caption{\textbf{Precision-Recall curve for the DeepArUco++ decoder on Shadow-ArUco dataset.} We can see the precision and recall values for different Hamming distance thresholds. \textit{Recall} is defined here as the fraction of markers with an ID found at a distance equal to or below the threshold; \textit{Precision} is defined as the fraction of markers with a correctly assigned ID w.r.t. the total of markers after filtering by that threshold.}
    \label{fig:decoder_pr}
\end{figure}

Using the method described in Section~\ref{ssec:decoder}, we have trained a simple marker decoding model based on the architecture from Fig.~\ref{fig:decoder}. The models have been trained on the cropped markers obtained from the Flying-ArUco v2 dataset, augmented by adding blur and color shift as we did for the corner refinement models in Section~\ref{sssec:results_regression}. 
In Fig.~\ref{fig:decoder_pr}, we can see the Precision-Recall curve obtained for the trained model on the markers from the Shadow-ArUco dataset (cropped and rectified from their ground-truth corner positions) at different maximum Hamming distances. From the curve, we can see how, even using a threshold of $9$, it is possible to correctly identify most markers (with a precision of $0.9903$); however, using a maximum distance of $3$, we can still achieve a recall of $0.9476$ while keeping a perfect precision.

%We can also use the marker decoding model to assess which corner refinement method should be used depending on our needs. 

\begin{figure}[t]
    \centering
    \includegraphics[width=0.75\textwidth]{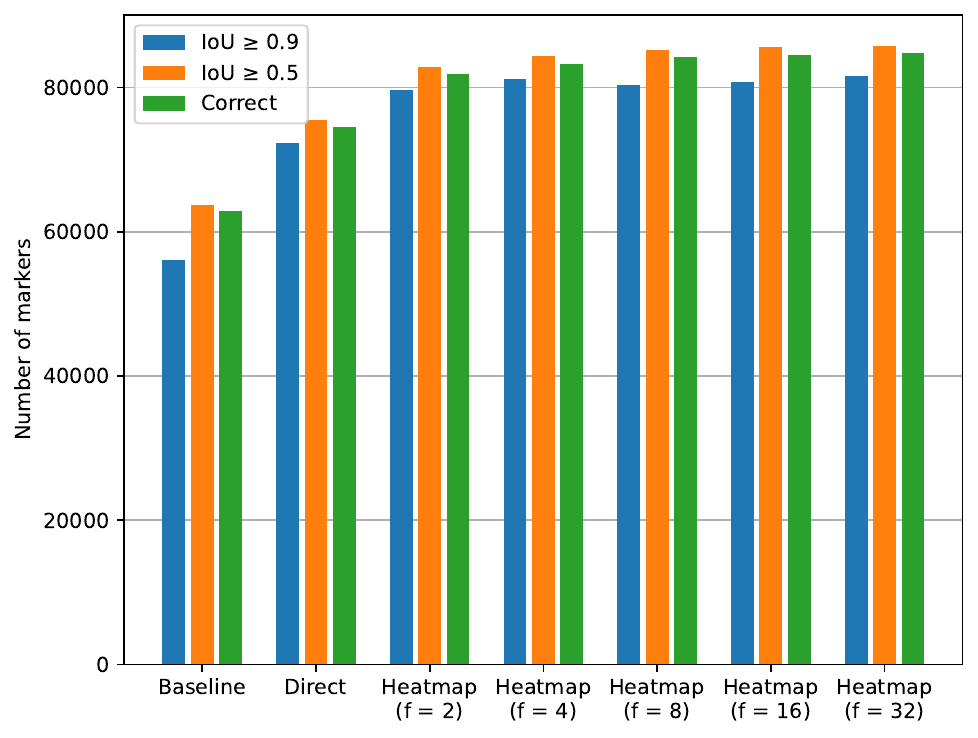}
    \caption{\textbf{Corner refinement precision vs. correctly decoded markers.} For each of the corner refinement models in Table~\ref{tab:regressor_comparison}, we compute the number of markers for which the IoU w.r.t. their ground-truth counterpart is greater or equal to each different threshold ($0.5$ and $0.9$, respectively), and the number of overall correctly decoded markers (without additional filtering).}
    \label{fig:thresholds}
\end{figure}

A relevant question regarding our pipeline is whether improvements in corner detection lead to improvements in marker identification or not, for instance, due to limitations in the approach to marker decoding. In Fig.~\ref{fig:thresholds}, we can see a comparison of the corner refinement precision of each method in Table~\ref{tab:regressor_comparison} in relation to the number of correctly identified markers obtained when applying the marker decoding model to the crops refined through such method. 
For each method, starting from the ground-truth marker locations (i.e., not using the detection methods described in Section~\ref{ssec:detector}), we crop and refine every marker in the Shadow-ArUco dataset from Section~\ref{ssec:shadowaruco}, and we obtain the number of markers for which the IoU with respect to the ground-truth counterpart is greater or equal than two different thresholds, namely $0.5$ and $0.9$; additionally, we apply the marker decoding model to the markers refined by each method (without additional filtering, i.e., no minimum IoU requirement has been enforced), and count the number of correctly decoded markers (i.e. the assigned label is the same as the ground-truth label). For both metrics, we have taken into account only markers with 4 detected corners.

From the results in Fig.~\ref{fig:thresholds}, it seems that the improvements achieved through the use of the heatmap-based approach do translate well to the number of correctly identified markers, with a near-perfect correlation (especially when considering the $0.5$ threshold). This suggests that the marker detection method does not hinder improvements in the rest of the pipeline. Additionally, we can use the obtained curves to determine which method to use at the corner refinement step: for instance, while the method which correctly identified the most markers is the heatmap-based with $f = 32$, the difference between it and the one with $f = 8$ is not that significant, with 84871 and 84312 respectively. However, the difference in inference time is somewhat noticeable (see Table~\ref{tab:regressor_comparison}).

%it seems that while the overall number of markers obtained by the heatmap-based models is slightly lower (the number of markers at the threshold of $0.5$ or more is slightly lower than for the direct regression approach), the overall precision in corner refinement increases (especially at the $0.9$ threshold the number of markers is substantially higher than for the direct-regression approach). At a glance, there could be a certain correlation between the number of markers at high IoU thresholds and the number of correctly identified markers; however, more experimentation is needed to assess that. For the purpose of this work, we will use the heatmap-based method with $f = 8$ as representative of the heatmap-based approach, as it leads to the highest amount of correctly identified markers after the marker decoding step.

% Comparison SOTA
\subsection{Ablation study.} \label{ssec:sota_comparison}

Finally, we will perform an ablation study comparing multiple configurations of the DeepArUco++ framework against each other. As our pipeline is composed of three clearly separated steps, and every step is required for providing the function of the complete system, our ablation study will focus on the multiple proposed combinations for marker detector and corner refinement models; the marker decoding model will be the same for every DeepArUco++ configuration. Also, we will compare against two currently state-of-the-art methods for the detection and decoding of fiducial markers, namely classic ArUco and DeepTag, and the baseline version of our method (DeepArUco), as presented in our previous work~\cite{berral2023ibpria}. While the baseline DeepArUco also uses a YOLO-based detector model, it was trained on a previous version of the Flying-ArUco dataset; also, it used a larger variant (\texttt{yolov8m}, instead of \texttt{yolov8s} or \texttt{yolov8n}). The tested DeepArUco++ configurations are the following:
%\begin{minipage}{\linewidth}
%\medskip
\begin{itemize}
    \item \texttt{yolov8n} + direct:
    \begin{itemize}
        \item Detector: Trained from \texttt{yolov8n}, using blur and color shift as offline data augmentation options.
        \item Refinement model: Direct regressor.
    \end{itemize}
    \item \texttt{yolov8n} + heatmap:
    \begin{itemize}
        \item Detector: Trained from \texttt{yolov8n}, using blur and color shift as offline data augmentation options.
        \item Refinement model: Heatmap-based, with $f = 8$.
    \end{itemize}
    \item \texttt{yolov8s} + direct:
    \begin{itemize}
        \item Detector: Trained from \texttt{yolov8s}, using blur and color shift as offline data augmentation options.
        \item Refinement model: Direct regressor.
    \end{itemize}
    \item \texttt{yolov8s} + heatmap:
    \begin{itemize}
        \item Detector: Trained from \texttt{yolov8s}, using blur and color shift as offline data augmentation options.
        \item Refinement model: Heatmap-based, with $f = 8$.
    \end{itemize}
\end{itemize}
%\medskip
%\end{minipage}

\begin{figure}[t]
    \centering
    \begin{subfigure}[t]{.49\textwidth}
        \centering
        \includegraphics[width=\textwidth]{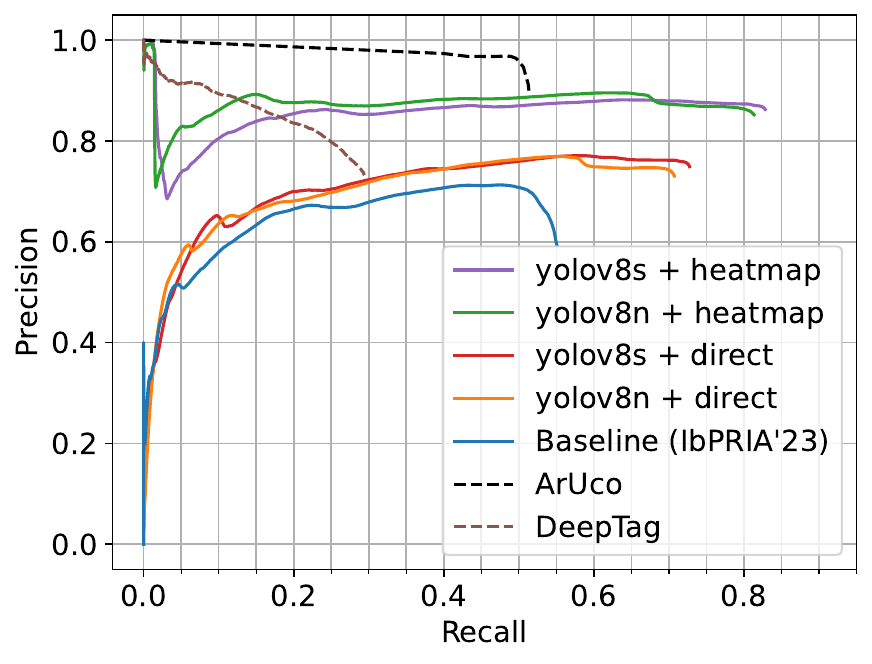}
        \caption{}
    \end{subfigure}
    \begin{subfigure}[t]{.49\textwidth}
        \centering
        \includegraphics[width=\textwidth]{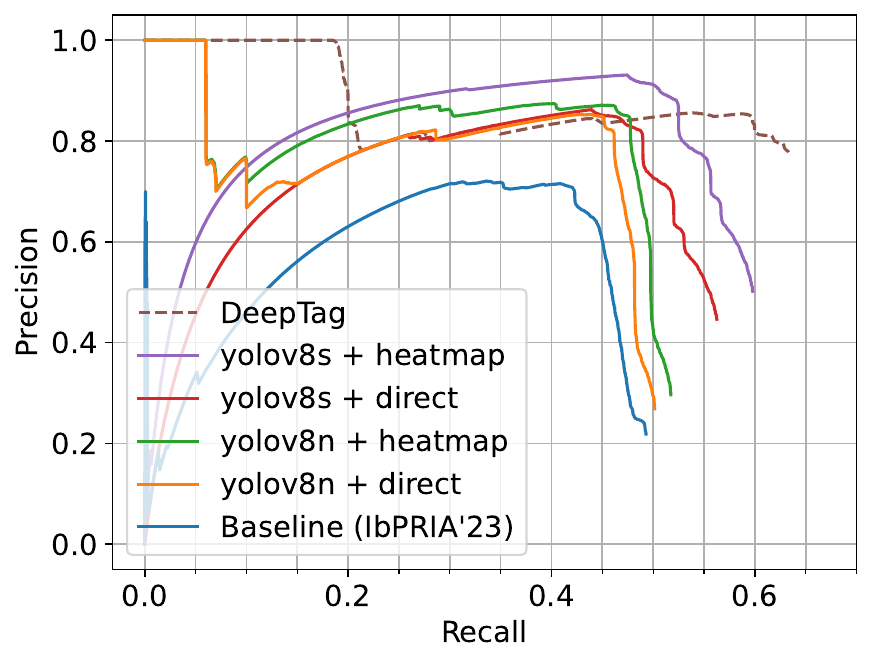}
        \caption{}
    \end{subfigure}
    \caption{\textbf{Comparison of DeepArUco++ detection capabilities against the state-of-the-art.} \textbf{(a)} Precision-Recall curve over the Shadow-ArUco dataset; \textbf{(b)} Precision-Recall over the DeepTag dataset. Legend reflects the considered method/setting. Best viewed in digital format.}
    \label{fig:sota_pr}
\end{figure}

%For each of the previous configurations, we have considered two options: using a value of $3$ as the Hamming distance threshold at the decoding step (i.e., discarding detected markers for which the distance to the nearest ID is greater than $3$), and not using any threshold (i.e., keeping every detected marker). 
No thresholding has been applied at the decoding step, and any markers for which less than four corners have been detected (with a maximum distance of 5 pixels to the closest ground-truth corner) have been removed. The Precision-Recall curve for each configuration over the Shadow-ArUco dataset appears in Fig.~\ref{fig:sota_pr}a.

As shown in Fig.~\ref{fig:sota_pr}a, there is a noticeable difference between the heatmap-based corner refinement model and the direct regression approach, with the former yielding significantly higher precision and recall values. For a given corner refinement model, the difference between using the \texttt{yolov8n} or the \texttt{yolov8s} detector is not that much. In any case, the detection capabilities of every tested DeepArUco++ configuration are far beyond those of ArUco, DeepTag, or even the baseline version of DeepArUco. However, it should be noted that in terms of precision, the classic ArUco is unmatched at the cost of a much-reduced recall. 
A numerical comparison of the different methods, considering the AUC value for the curves in Fig.~\ref{fig:sota_pr}a, the mean error at the corners and the percentage of detected markers at different levels of restrictiveness appears in Table~\ref{tab:sota_comparison}.

\begin{table}[t]
    \centering
    \caption{\textbf{Quantitative comparison of DeepArUco++ detection capabilities against the state-of-the-art, on Shadow-ArUco dataset.} 
    \textit{Matched B.B.} stands for the percentage of markers detected, discarding markers with less than 4 detected corners at the output of the corner refinement step; \textit{Corners filtered} stands for the percentage of detected markers that remain after discarding every corner for which the distance to the closest ground-truth corner is higher than 5 pixels, and keeping only markers with 4 remaining corners; \textit{Corners + ID.} stands for the percentage of markers after filtering corners for which the predicted ID. is correct; \textit{Only ID.} stands for the percentage of detected markers for which a correct ID. has been predicted, but less than 4 corners remain after filtering; \textit{Corner error} stands for the mean Euclidean distance between a predicted corner and its nearest ground-truth corner (in pixels). For metrics noted with $\downarrow$, lower is better; for $\uparrow$, higher is better.}

    \small
    \begin{tabular}{|c|c|ccc|c|c|}
    \cline{3-6}
    \multicolumn{2}{c|}{} & \multicolumn{4}{c|}{\% of ground-truth markers} & \multicolumn{1}{c}{}\\
    \hline
    \multirow{2}{*}{\makecell[c]{\textbf{Method}}} & \multirow{2}{*}{\makecell[c]{\textbf{AUC $\uparrow$}}} & \multirow{2}{*}{\makecell[c]{\textbf{Matched}\\\textbf{B.B. $\uparrow$}}} & \multirow{2}{*}{\makecell[c]{\textbf{Corners}\\\textbf{filtered $\uparrow$}}} & \multirow{2}{*}{\makecell[c]{\textbf{Corners}\\\textbf{+ ID. $\uparrow$}}} & \multirow{2}{*}{\makecell[c]{\textbf{Only}\\\textbf{ID.}}} & \multirow{2}{*}{\makecell[c]{\textbf{Corner}\\\textbf{error $\downarrow$}}}\\
    & & & & & &\\
    \hline
    \hline
    \multirow{2}{*}{\makecell[c]{yolov8n +\\direct}} & \multirow{2}{*}{\makecell[c]{0.4930}} & \multirow{2}{*}{\makecell[c]{94.80\%}} & \multirow{2}{*}{\makecell[c]{70.74\%}} & \multirow{2}{*}{\makecell[c]{69.64\%}} & \multirow{2}{*}{\makecell[c]{18.56\%}} & \multirow{2}{*}{\makecell[c]{1.94 $\pm$\\ 0.95}}\\
    & & & & & &\\
    \hline
    \multirow{2}{*}{\makecell[c]{yolov8n +\\heatmap}} & \multirow{2}{*}{\makecell[c]{\textbf{0.7120}}} & \multirow{2}{*}{\makecell[c]{94.24\%}} & \multirow{2}{*}{\makecell[c]{81.34\%}} & \multirow{2}{*}{\makecell[c]{79.68\%}} & \multirow{2}{*}{\makecell[c]{8.75\%}} & \multirow{2}{*}{\makecell[c]{2.11 $\pm$\\ 0.83}}\\
    & & & & & &\\
    \hline
    \hline
    \multirow{2}{*}{\makecell[c]{yolov8s +\\direct}} & \multirow{2}{*}{\makecell[c]{0.5107}} & \multirow{2}{*}{\makecell[c]{\textbf{95.82\%}}} & \multirow{2}{*}{\makecell[c]{72.75\%}} & \multirow{2}{*}{\makecell[c]{71.38\%}} & \multirow{2}{*}{\makecell[c]{17.70\%}} & \multirow{2}{*}{\makecell[c]{1.89 $\pm$\\ 0.95}}\\
    & & & & & &\\
    \hline
    \multirow{2}{*}{\makecell[c]{yolov8s +\\heatmap}} & \multirow{2}{*}{\makecell[c]{0.7094}} & \multirow{2}{*}{\makecell[c]{95.11\%}} & \multirow{2}{*}{\makecell[c]{\textbf{82.83\%}}} & \multirow{2}{*}{\makecell[c]{\textbf{80.90\%}}} & \multirow{2}{*}{\makecell[c]{8.26\%}} & \multirow{2}{*}{\makecell[c]{2.04 $\pm$\\ 0.83}}\\
    & & & & & &\\
    \hline
    \hline
    \multirow{2}{*}{\makecell[c]{Baseline\\(IbPRIA'23)}} & \multirow{2}{*}{\makecell[c]{0.3530}} & \multirow{2}{*}{\makecell[c]{82.35\%}} & \multirow{2}{*}{\makecell[c]{55.06\%}} & \multirow{2}{*}{\makecell[c]{54.78\%}} & \multirow{2}{*}{\makecell[c]{21.99\%}} & \multirow{2}{*}{\makecell[c]{1.77 $\pm$\\ 1.10}}\\
    & & & & & &\\
    \hline
    \hline
    \multirow{2}{*}{\makecell[c]{ArUco}} & \multirow{2}{*}{\makecell[c]{0.5047}} & \multirow{2}{*}{\makecell[c]{53.25\%}} & \multirow{2}{*}{\makecell[c]{51.41\%}} & \multirow{2}{*}{\makecell[c]{43.60\%}} & \multirow{2}{*}{\makecell[c]{1.40\%}} & \multirow{2}{*}{\makecell[c]{\textbf{0.72 $\pm$}\\ \textbf{0.51}}}\\
    & & & & & &\\
    \hline
    \hline
    \multirow{2}{*}{\makecell[c]{Deeptag}} & \multirow{2}{*}{\makecell[c]{0.2537}} & \multirow{2}{*}{\makecell[c]{39.77\%}} & \multirow{2}{*}{\makecell[c]{29.38\%}} & \multirow{2}{*}{\makecell[c]{29.31\%}} & \multirow{2}{*}{\makecell[c]{10.13\%}} & \multirow{2}{*}{\makecell[c]{1.40 $\pm$\\ 0.93}}\\
    & & & & & &\\
    \hline
    \end{tabular}
    \label{tab:sota_comparison}
\end{table}

From the comparison on Table~\ref{tab:sota_comparison}, we can conclude the following: if we only consider the Intersection-over-Union (IoU) between the detected markers and the ground-truth markers (considering every predicted corner as valid, and only considering markers with $4$ detected corners) the percentage of found markers (with respect to the complete Shadow-ArUco dataset) is similar for every DeepArUco++ configuration. 
This seems to indicate that in this dataset, the penalty when using the \texttt{yolov8n}-based detector versus the \texttt{yolov8s}-based one is not that much. However, when discarding corners further than $5$ pixels from their closest ground-truth corner (and keeping only markers retaining the four corners), differences arise between the heatmap-based and the direct corner regression approach, with the former resulting in a much higher yield.
For every method (with the exception of ArUco), it seems that almost every marker with $4$ corners found can be correctly decoded. Also, it is worth noticing that in the case of DeepTag, almost every detected marker can be decoded (counting every marker with $4$ detected corners before filtering and a correctly assigned ID., for both ``Corners + ID.'' and ``Only ID.'' columns). 
Regarding per-corner error, the results are also similar for every DeepArUco++ configuration, with marginally worse values for the configurations using the heatmap-based corner refinement model. However, this could stem from the fact that way more corners are being detected with this approach, possibly including some extreme cases not detected by the direct regression approach. The smallest per-corner error is achieved by the classic ArUco implementation; this could be the reason behind the almost perfect match between the number of detected markers and the number of markers with 4 corners remaining after filtering.

\textbf{Evaluation on DeepTag dataset.} We have also tested our methods using a different dataset: in this case, we have considered the dataset used to develop DeepTag. A comparison of the detection capabilities of the different DeepArUco++ configurations against DeepTag on the DeepTag dataset (the portion of the \texttt{general} dataset containing ArUco markers) appears in Fig.~\ref{fig:sota_pr}b. 

\begin{figure}[t]
    \centering
    \includegraphics[width=\textwidth]{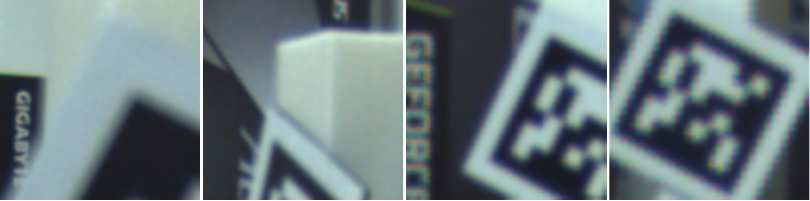}
    \vspace{0.5mm}
    \caption{\textbf{Detail of some samples from the DeepTag ArUco dataset.} Note the presence of a certain overall ``blockiness'' and color noise around the edges of the markers and other elements in the scene. Best viewed in digital format.}
    \label{fig:deeptag_detail}
\end{figure}

The results in Fig.~\ref{fig:sota_pr}b indicate that, under our test conditions, our method remains competitive on the DeepTag dataset, showing only a slight reduction in precision and recall when compared to the DeepTag method (when considering the \texttt{yolov8s} + heatmap configuration). Please note that we did not implement any specific measures to address particular characteristics of the DeepTag dataset, namely the noise and overall ``blockiness'' of the images (see Fig.~\ref{fig:deeptag_detail}), which may have reduced the detection capabilities of our models. A numerical comparison of the different methods on this new dataset appears in Table~\ref{tab:sota_comparison_deeptag}.

\begin{table}[t]
    \centering
    \caption{\textbf{Quantitative comparison of DeepArUco++ detection capabilities against the state-of-the-art, on DeepTag dataset.} \textit{Matched B.B.} stands for the percentage of markers detected, discarding markers with less than 4 detected corners at the output of the corner refinement step; \textit{Corners filtered} stands for the percentage of detected markers that remain after discarding every corner for which the distance to the closest ground-truth corner is higher than 5 pixels, and keeping only markers with 4 remaining corners; \textit{Corners + ID.} stands for the percentage of markers after filtering corners for which the predicted ID. is correct; \textit{Only ID.} stands for the percentage of detected markers for which a correct ID. has been predicted, but less than 4 corners remain after filtering; \textit{Corner error} stands for the mean Euclidean distance between a predicted corner and its nearest ground-truth corner (in pixels). For metrics noted with $\downarrow$, lower is better; for $\uparrow$, higher is better.}

    \small
    \begin{tabular}{|c|c|ccc|c|c|}
    \cline{3-6}
    \multicolumn{2}{c|}{} & \multicolumn{4}{c|}{\% of ground-truth markers} & \multicolumn{1}{c}{}\\
    \hline
    \multirow{2}{*}{\makecell[c]{\textbf{Method}}} & \multirow{2}{*}{\makecell[c]{\textbf{AUC $\uparrow$}}} & \multirow{2}{*}{\makecell[c]{\textbf{Matched}\\\textbf{B.B. $\uparrow$}}} & \multirow{2}{*}{\makecell[c]{\textbf{Corners}\\\textbf{filtered $\uparrow$}}} & \multirow{2}{*}{\makecell[c]{\textbf{Corners}\\\textbf{+ ID. $\uparrow$}}} & \multirow{2}{*}{\makecell[c]{\textbf{Only}\\\textbf{ID.}}} & \multirow{2}{*}{\makecell[c]{\textbf{Corner}\\\textbf{error $\downarrow$}}}\\
    & & & & & &\\
    \hline
    \hline
    \multirow{2}{*}{\makecell[c]{yolov8n +\\direct}} & \multirow{2}{*}{\makecell[c]{0.3972}} & \multirow{2}{*}{\makecell[c]{59.37\%}} & \multirow{2}{*}{\makecell[c]{50.15\%}} & \multirow{2}{*}{\makecell[c]{50.14\%}} & \multirow{2}{*}{\makecell[c]{8.82\%}} & \multirow{2}{*}{\makecell[c]{2.18 $\pm$\\ 0.82}}\\
    & & & & & &\\
    \hline
    \multirow{2}{*}{\makecell[c]{yolov8n +\\heatmap}} & \multirow{2}{*}{\makecell[c]{0.4288}} & \multirow{2}{*}{\makecell[c]{59.37\%}} & \multirow{2}{*}{\makecell[c]{51.74\%}} & \multirow{2}{*}{\makecell[c]{43.28\%}} & \multirow{2}{*}{\makecell[c]{7.23\%}} & \multirow{2}{*}{\makecell[c]{2.94 $\pm$\\ 0.82}}\\
    & & & & & &\\
    \hline
    \hline
    \multirow{2}{*}{\makecell[c]{yolov8s +\\direct}} & \multirow{2}{*}{\makecell[c]{0.3967}} & \multirow{2}{*}{\makecell[c]{65.23\%}} & \multirow{2}{*}{\makecell[c]{56.26\%}} & \multirow{2}{*}{\makecell[c]{55.22\%}} & \multirow{2}{*}{\makecell[c]{7.02\%}} & \multirow{2}{*}{\makecell[c]{\textbf{1.97 $\pm$}\\ \textbf{0.84}}}\\
    & & & & & &\\
    \hline
    \multirow{2}{*}{\makecell[c]{yolov8s +\\heatmap}} & \multirow{2}{*}{\makecell[c]{\textbf{0.4785}}} & \multirow{2}{*}{\makecell[c]{65.23\%}} & \multirow{2}{*}{\makecell[c]{59.79\%}} & \multirow{2}{*}{\makecell[c]{57.64\%}} & \multirow{2}{*}{\makecell[c]{3.60\%}} & \multirow{2}{*}{\makecell[c]{2.62 $\pm$\\ 0.81}}\\
    & & & & & &\\
    \hline
    \hline
    \multirow{2}{*}{\makecell[c]{Baseline\\(IbPRIA'23)}} & \multirow{2}{*}{\makecell[c]{0.2760}} & \multirow{2}{*}{\makecell[c]{64.00\%}} & \multirow{2}{*}{\makecell[c]{49.30\%}} & \multirow{2}{*}{\makecell[c]{49.30\%}} & \multirow{2}{*}{\makecell[c]{14.70\%}} & \multirow{2}{*}{\makecell[c]{2.70 $\pm$\\ 0.77}}\\
    & & & & & &\\
    \hline
    \hline
    \multirow{2}{*}{\makecell[c]{DeepTag}} & \multirow{2}{*}{\makecell[c]{0.5583}} & \multirow{2}{*}{\makecell[c]{\textbf{81.72\%}}} & \multirow{2}{*}{\makecell[c]{\textbf{63.36\%}}} & \multirow{2}{*}{\makecell[c]{\textbf{63.36\%}}} & \multirow{2}{*}{\makecell[c]{18.36\%}} & \multirow{2}{*}{\makecell[c]{2.49 $\pm$\\1.07}}\\
    & & & & & &\\
    \hline
    \end{tabular}
    \label{tab:sota_comparison_deeptag}
\end{table}

The data in Table~\ref{tab:sota_comparison_deeptag} leads us to a similar conclusion to that obtained from Fig.~\ref{fig:sota_pr}b. When considering only marker detection, it seems like DeepTag is able to recover the most markers. 
However, the difference becomes much smaller when filtering corners far from their nearest ground-truth, especially when compared against the configurations using the \texttt{yolov8s}-based marker detector. On a curious note, it seems that both our baseline method and DeepTag achieve perfect precision when identifying markers (counting every marker for both “Corners + ID.” and “Only ID.” columns), regardless of the corner location accuracy. 
Regarding per-corner error, it seems that the configurations using the direct regression approach in the corner refinement step have the smallest error, with the rest of the methods presenting similar performance. Please note that for this dataset, precise marker locations were not provided, and we had to annotate the corner positions manually.

%Significant differences exist between the datasets used to develop our methods and those used to develop DeepTag. However, we can highlight that our methods can achieve a smaller average per-corner error with less variability. Please note that for this dataset, precise marker locations were not provided, and we had to annotate the corner positions manually.

\textbf{Cross-validation measurements.} As an additional assessment of the performance of our methods, we have performed a 5-fold cross-validation, training and testing over portions of the Flying-ArUco v2 dataset. To this effect, we have split the Flying-ArUco v2 dataset into $5$ different parts  after augmenting its samples by using blur and color shift. Then, we trained our entire pipeline over $4$ of such parts and tested the remaining. We repeated this process five times for each configuration. The aggregate quantitative metrics appear in Table \ref{tab:crossvalidation}.

\begin{table}[t]
    \centering
    \caption{\textbf{Quantitative comparison of multiple DeepArUco++ configurations, using 5-fold cross-validation.} \textit{Matched B.B.} stands for the percentage of markers detected, discarding markers with less than 4 detected corners at the output of the corner refinement step; \textit{Corners filtered} stands for the percentage of detected markers that remain after discarding every corner for which the distance to the closest ground-truth corner is higher than 5 pixels, and keeping only markers with 4 remaining corners; \textit{Corners + ID.} stands for the percentage of markers after filtering corners for which the predicted ID. is correct; \textit{Only ID.} stands for the percentage of detected markers for which a correct ID. has been predicted, but less than 4 corners remain after filtering; \textit{Corner error} stands for the mean Euclidean distance between a predicted corner and its nearest ground-truth corner (in pixels). For metrics noted with $\downarrow$, lower is better; for $\uparrow$, higher is better.}
    \small
    \begin{tabular}{|c|c|ccc|c|c|}
    \cline{3-6}
    \multicolumn{2}{c|}{} & \multicolumn{4}{c|}{\% of ground-truth markers} & \multicolumn{1}{c}{}\\
    \hline
    \multirow{2}{*}{\makecell[c]{\textbf{Method}}} & \multirow{2}{*}{\makecell[c]{\textbf{AUC $\uparrow$}}} & \multirow{2}{*}{\makecell[c]{\textbf{Matched}\\\textbf{B.B. $\uparrow$}}} & \multirow{2}{*}{\makecell[c]{\textbf{Corners}\\\textbf{filtered $\uparrow$}}} & \multirow{2}{*}{\makecell[c]{\textbf{Corners}\\\textbf{+ ID. $\uparrow$}}} & \multirow{2}{*}{\makecell[c]{\textbf{Only}\\\textbf{ID.}}} & \multirow{2}{*}{\makecell[c]{\textbf{Corner}\\\textbf{error $\downarrow$}}}\\
    & & & & & &\\
    \hline
    \hline
    \multirow{2}{*}{\makecell[c]{yolov8n +\\direct}} & \multirow{2}{*}{\makecell[c]{0.9182 $\pm$\\ 0.0484}} & \multirow{2}{*}{\makecell[c]{99.44 $\pm$\\ 0.18\%}} & \multirow{2}{*}{\makecell[c]{96.07 $\pm$\\ 1.81\%}} & \multirow{2}{*}{\makecell[c]{72.81 $\pm$\\ 0.90\%}} & \multirow{2}{*}{\makecell[c]{1.75 $\pm$\\ 1.27\%}} & \multirow{2}{*}{\makecell[c]{1.47 $\pm$\\ 0.65}}\\
    & & & & & &\\
    \hline
    \multirow{2}{*}{\makecell[c]{yolov8n +\\heatmap}} & \multirow{2}{*}{\makecell[c]{0.9729 $\pm$\\ 0.0050}} & \multirow{2}{*}{\makecell[c]{98.22 $\pm$\\ 0.38\%}} & \multirow{2}{*}{\makecell[c]{97.66 $\pm$\\ 0.43\%}} & \multirow{2}{*}{\makecell[c]{\textbf{74.71 $\pm$}\\ \textbf{3.09\%}}} & \multirow{2}{*}{\makecell[c]{0.23 $\pm$\\ 0.08\%}} & \multirow{2}{*}{\makecell[c]{1.65 $\pm$\\ 0.60}}\\
    & & & & & &\\
    \hline
    \hline
    \multirow{2}{*}{\makecell[c]{yolov8s +\\direct}} & \multirow{2}{*}{\makecell[c]{0.9170 $\pm$\\ 0.0521}} & \multirow{2}{*}{\makecell[c]{\textbf{99.57 $\pm$}\\ \textbf{0.08\%}}} & \multirow{2}{*}{\makecell[c]{96.23 $\pm$\\ 1.75\%}} & \multirow{2}{*}{\makecell[c]{72.74 $\pm$\\ 0.67\%}} & \multirow{2}{*}{\makecell[c]{1.76 $\pm$\\ 1.24\%}} & \multirow{2}{*}{\makecell[c]{\textbf{1.47 $\pm$}\\ \textbf{0.63}}}\\
    & & & & & &\\
    \hline
    \multirow{2}{*}{\makecell[c]{yolov8s +\\heatmap}} & \multirow{2}{*}{\makecell[c]{\textbf{0.9742 $\pm$}\\ \textbf{0.0042}}} & \multirow{2}{*}{\makecell[c]{98.28 $\pm$\\ 0.31\%}} & \multirow{2}{*}{\makecell[c]{\textbf{97.73 $\pm$}\\ \textbf{0.34\%}}} & \multirow{2}{*}{\makecell[c]{74.69 $\pm$\\ 3.09\%}} & \multirow{2}{*}{\makecell[c]{0.21 $\pm$\\ 0.08\%}} & \multirow{2}{*}{\makecell[c]{1.65 $\pm$\\ 0.59}}\\
    & & & & & &\\
    \hline
    \end{tabular}
    \label{tab:crossvalidation}
\end{table}

From the results in \ref{tab:crossvalidation}, we can see how for a given detector size (\texttt{yolov8n} or \texttt{yolov8s}) the choice of corner refinement model has the most impact in terms of both precise location and decoding performance, with the heatmap-based corner refinement model yielding the best results overall. Again, this would indicate that the heatmap-based approach manages to recover the most corners (even if the measured error \textit{for recovered corners} is lower when using the direct-regression approach). On the other hand, the choice of marker detector size has not that much of an impact on this dataset.

\subsection{Throughput comparison.} 

Finally, we can compare the different methods from a throughput standpoint. We have considered only two configurations, \texttt{yolov8n} detector with direct corner regression in the refinement step, and \texttt{yolov8s} detector with heatmap-based corner refinement. For both configurations, we have used the same marker decoding model. For DeepArUco++ configurations, we have timed the three stages: marker detection, corner regression (including the operations required to extract the final corner values when using a heatmap-based model), and marker decoding (including the time required to identify the marker ID). Additionally, we measure the time required for processing each frame (without including the time spent loading the image or writing the results). For DeepTag and classic ArUco, only the total time per frame has been measured. Results of this comparison appear in Table~\ref{tab:time_breakdown}.

\begin{table}[t]
    \centering
    \caption{\textbf{Time breakdown of the studied methods.} \textit{Det. time} stands for the time spent per frame at the marker detection step (if applicable); \textit{Ref. time} stands for the time spent at the corner refinement step; \textit{Dec. time} stands for the time spent at the marker decoding step; \textit{Total time} stands for the total time per frame (i.e. the time spent on every step, plus any extra time spent on operations not belonging to any step). Times are measured in milliseconds. Picture resolution is $800 \times 600$, with an average of $10.90 \pm 1.33$ markers per image. For metrics noted with $\downarrow$, lower is better; for $\uparrow$, higher is better.}
    \scriptsize
    \begin{tabular}{|c|c|c|c|c|c|c|}
        \hline
        \multirow{2}{*}{\makecell[c]{\textbf{Method}}} & \multirow{2}{*}{\makecell[c]{\textbf{Det. time $\downarrow$}}} & \multirow{2}{*}{\makecell[c]{\textbf{Ref. time $\downarrow$}}} & \multirow{2}{*}{\makecell[c]{\textbf{Dec. time $\downarrow$}}} & \multirow{2}{*}{\makecell[c]{\textbf{Total time $\downarrow$}}} & \multirow{2}{*}{\makecell[c]{\textbf{FPS $\uparrow$}}}\\
        & & & & &\\
        \hline
        \hline
        \multirow{3}{*}{\makecell[c]{\texttt{yolov8n} +\\ direct}} & \multirow{3}{*}{\makecell[c]{\textbf{8.05 $\pm$}\\ \textbf{7.63}}} & \multirow{3}{*}{\makecell[c]{3.41 $\pm$\\ 5.39}} & \multirow{3}{*}{\makecell[c]{13.37 $\pm$\\ 2.09}} & \multirow{3}{*}{\makecell[c]{26.87 $\pm$\\ 12.98}} & \multirow{3}{*}{\makecell[c]{37.78 $\pm$\\ 3.21}}\\
        & & & & &\\
        & & & & &\\
        \hline
        \multirow{3}{*}{\makecell[c]{\texttt{yolov8s} +\\ heatmap}} & \multirow{3}{*}{\makecell[c]{8.08 $\pm$\\ 7.91}} & \multirow{3}{*}{\makecell[c]{6.94 $\pm$\\ 4.90}} & \multirow{3}{*}{\makecell[c]{\textbf{13.00 $\pm$}\\ \textbf{1.42}}} & \multirow{3}{*}{\makecell[c]{29.86 $\pm$\\ 12.55}} & \multirow{3}{*}{\makecell[c]{33.88 $\pm$\\ 2.34}}\\
        & & & & &\\
        & & & & &\\
        \hline
        \hline
        \multirow{3}{*}{\makecell[c]{Baseline\\ (IbPRIA'23)}} & \multirow{3}{*}{\makecell[c]{9.77 $\pm$\\ 8.80}} & \multirow{3}{*}{\makecell[c]{\textbf{3.28 $\pm$}\\ \textbf{5.78}}} & \multirow{3}{*}{\makecell[c]{13.44 $\pm$\\ 3.43}} & \multirow{3}{*}{\makecell[c]{28.18 $\pm$\\ 14.64}} & \multirow{3}{*}{\makecell[c]{36.46 $\pm$\\ 4.73}}\\
        & & & & &\\
        & & & & &\\
        \hline
        \hline
        \multirow{3}{*}{\makecell[c]{Classic\\ ArUco}} & \multirow{3}{*}{\makecell[c]{-}} & \multirow{3}{*}{\makecell[c]{-}} & \multirow{3}{*}{\makecell[c]{-}} & \multirow{3}{*}{\makecell[c]{\textbf{8.57 $\pm$}\\ \textbf{2.79}}} & \multirow{3}{*}{\makecell[c]{\textbf{130.52 $\pm$}\\ \textbf{44.62}}}\\
        & & & & &\\
        & & & & &\\
        \hline
        \hline
        \multirow{3}{*}{\makecell[c]{DeepTag}} & \multirow{3}{*}{\makecell[c]{-}} & \multirow{3}{*}{\makecell[c]{-}} & \multirow{3}{*}{\makecell[c]{-}} & \multirow{3}{*}{\makecell[c]{142.76 $\pm$\\ 114.16}} & \multirow{3}{*}{\makecell[c]{28.66 $\pm$\\ 41.45}}\\
        & & & & &\\
        & & & & &\\
        \hline
    \end{tabular}
    \label{tab:time_breakdown}
\end{table}

From data in Table~\ref{tab:time_breakdown}, we can see how our methods achieve a significant throughput, with average per-frame times below $30$ milliseconds. The high variance in time metrics reflects the challenges of accurately measuring short runtimes, which are affected by factors such as I/O activity and CPU load. The following conclusions can be drawn: first, there is little difference in inference times between using a detector based in \texttt{yolov8n} versus a detector based in \texttt{yolov8s}. The difference between \texttt{yolov8s} and \texttt{yolov8m} (as in the baseline DeepArUco) is more noticeable. On the corner refinement step, we can see a clear advantage in throughput when using the direct coordinate regression approach (as in the \texttt{yolov8n} + direct configuration, or the baseline DeepArUco) versus using the heatmap-based method. Finally, we can note how our methods outperform DeepTag from a throughput standpoint, with higher average FPS and way less variability.
%with total inference times smaller by $\sim6\times$.

\subsection{Working examples and limitations.} \label{ssec:qualresults}

\begin{figure}[t]
    \centering
    \begin{subfigure}[t]{.32\textwidth}
        \centering
        \includegraphics[width=\textwidth]{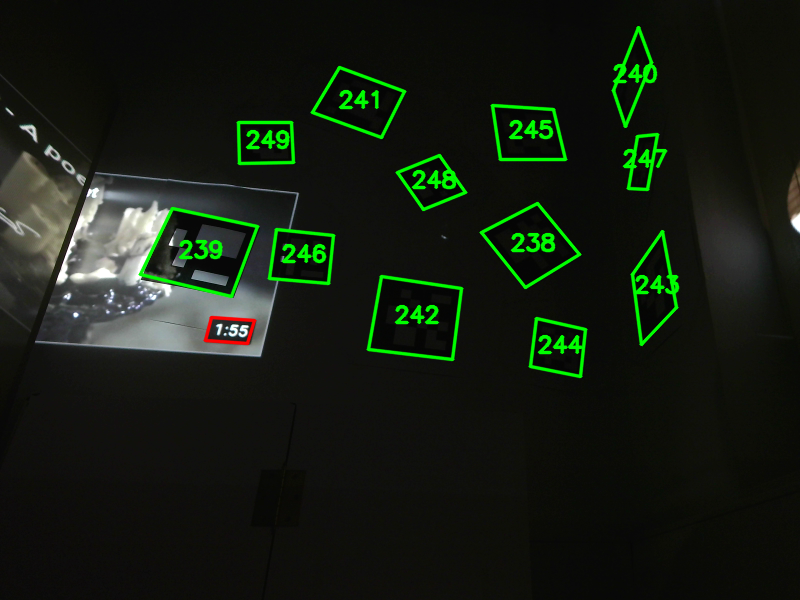}
        \caption{}
    \end{subfigure}
    \begin{subfigure}[t]{.32\textwidth}
        \centering
        \includegraphics[width=\textwidth]{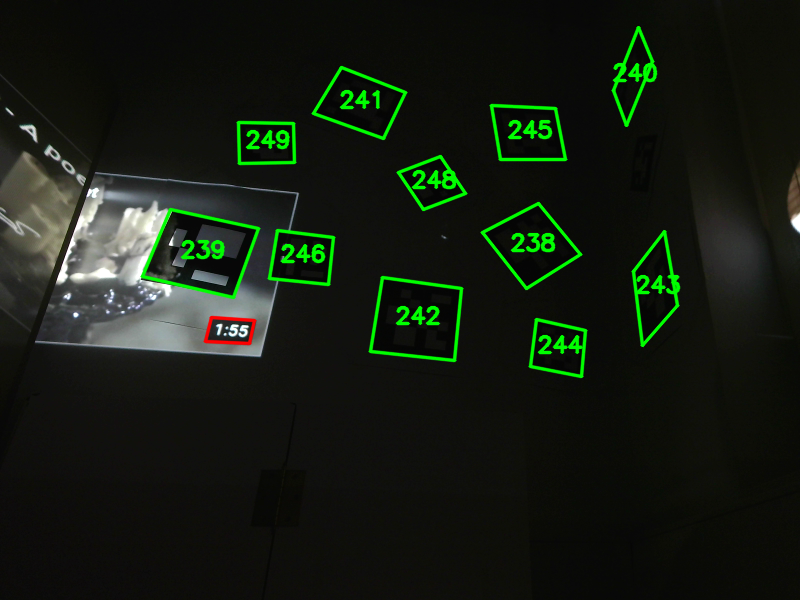}
        \caption{}
    \end{subfigure}
    \begin{subfigure}[t]{.32\textwidth}
        \centering
        \includegraphics[width=\textwidth]{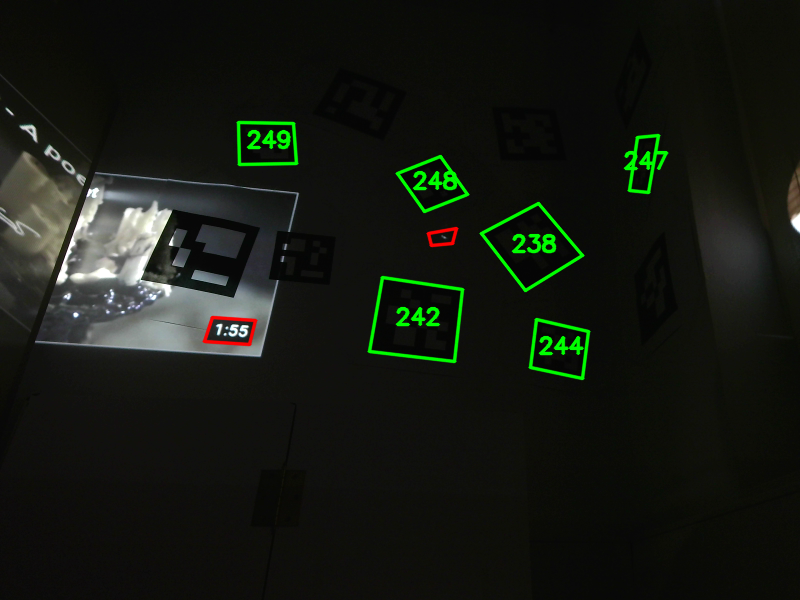}
        \caption{}
    \end{subfigure}
    \caption{\textbf{Qualitative evaluation against SOTA methods, on Shadow-ArUco dataset on poor lighting conditions.} We have considered \texttt{yolov8s} + heatmap (a), \texttt{yolov8n} + direct (b), and classic ArUco (c). Red outlines indicate false positive detections. Red IDs indicate wrong identifications. (Best viewed in digital format)}
    \label{fig:demo_dark}
\end{figure}

In Fig.~\ref{fig:demo_dark}, we can see how both \texttt{yolov8s} + heatmap (a) and \texttt{yolov8n} + direct (b) manage to outperform classic ArUco (c) on the task of detecting ArUco markers in pitch-black conditions. This scene does not seem to pose any difficulties, with both configurations performing well enough, with \texttt{yolov8s} + heatmap recovering an extra marker over \texttt{yolov8n} + direct. 
Please note that false detections may occur (e.g., the clock at the left of the scene). However, our method can discard most of these false detections by careful thresholding. 
On the other hand, DeepTag did not manage to recover any markers in this scene. Overall, from this image we can conclude that while difficult for the human eye, detecting and decoding markers in a darkened scene is feasible, provided the scene is uniformly dim.

\begin{figure}[t]
    \centering
    \begin{subfigure}[t]{.49\textwidth}
        \centering
        \includegraphics[width=\textwidth]{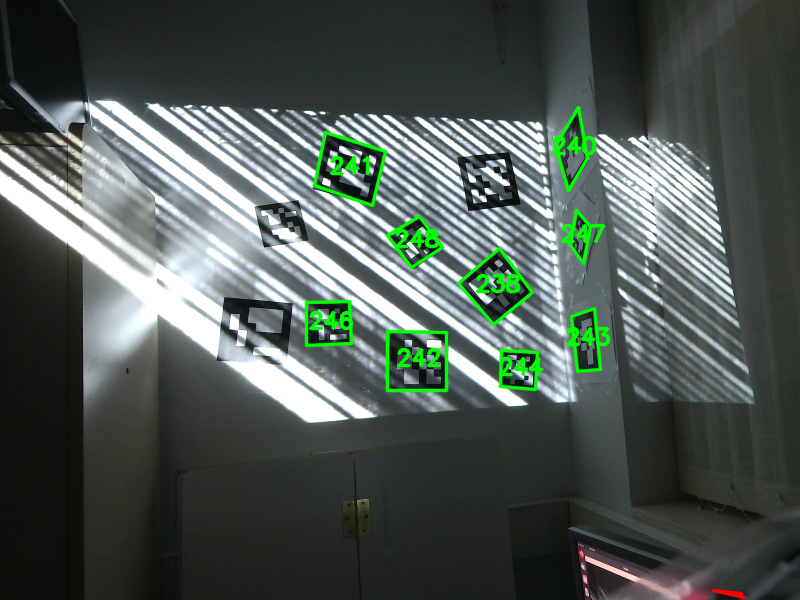}
        \caption{}
    \end{subfigure}
    \begin{subfigure}[t]{.49\textwidth}
        \centering
        \includegraphics[width=\textwidth]{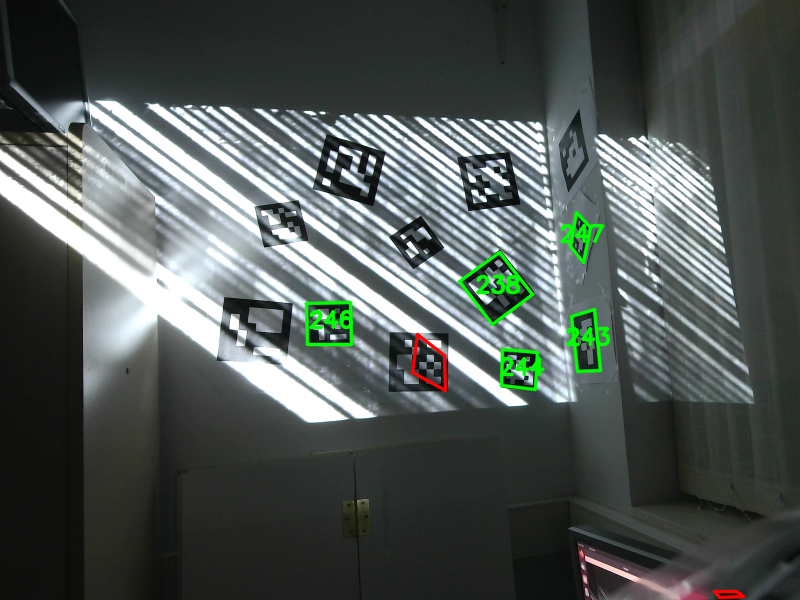}
        \caption{}
    \end{subfigure}
    \begin{subfigure}[t]{.49\textwidth}
        \centering
        \includegraphics[width=\textwidth]{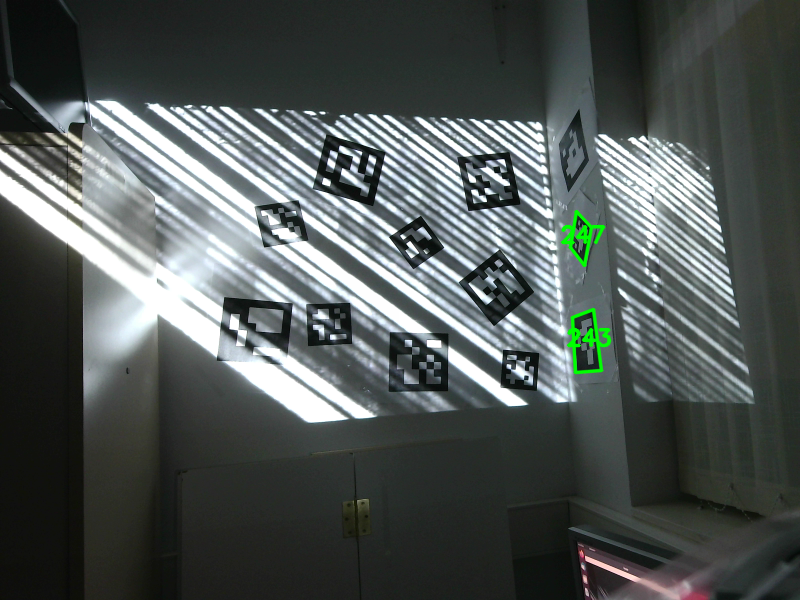}
        \caption{}
    \end{subfigure}
    \begin{subfigure}[t]{.49\textwidth}
        \centering
        \includegraphics[width=\textwidth]{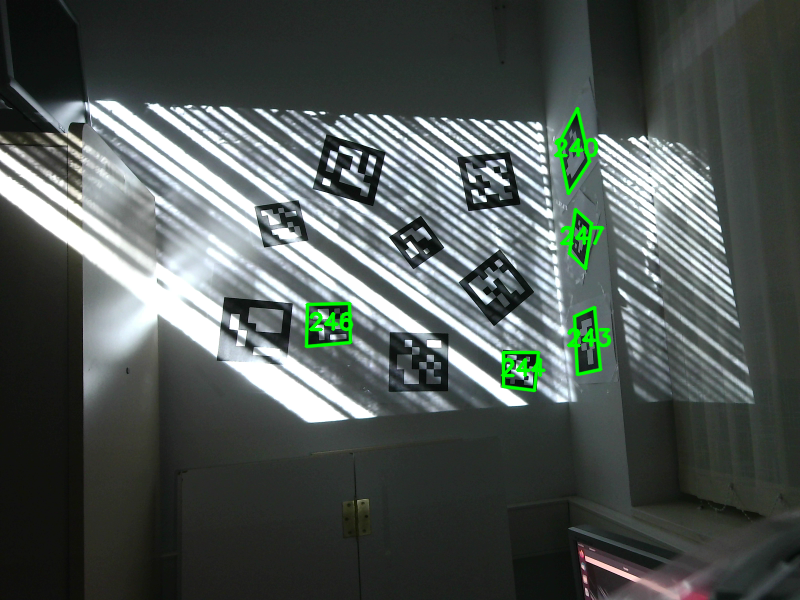}
        \caption{}
    \end{subfigure}
    \caption{\textbf{Qualitative evaluation against SOTA methods, on Shadow-ArUco dataset with complex lighting patterns.} We have considered \texttt{yolov8s} + heatmap (a), \texttt{yolov8n} + direct (b), classic ArUco (c), and DeepTag (d). Red outlines indicate false positive detections. Red IDs indicate wrong identifications. (Best viewed in digital format)}
    \label{fig:demo_stripes}
\end{figure}

While detecting markers on a uniformly darkened scene may seem difficult, detecting those in a scene with complex lighting patterns poses a more significant challenge, as they can mess with the edges of the markers, making the detection difficult. In Fig.~\ref{fig:demo_stripes}, we can see how both variants of DeepArUco++ (a and b) manage to find and correctly decode more markers compared to both ArUco (c) and DeepTag (d). 
In this case, we can also see how the heatmap-based corner refinement model on \texttt{yolov8s} + heatmap (a) can outperform the direct regression method in \texttt{yolov8n} + direct (b), as exemplified by the marker with ID 242 in (a). The improvement in corner detection leads to a better crop, enabling the correct identification of the marker. Both classic ArUco (c) and DeepTag (d) fail to detect most markers in the scene.
%Please note that while \texttt{yolov8s} + heatmap manages to recover a larger area of the marker, it is still not enough to recover a correct ID. While classic ArUco (c) and DeepTag (d) do not introduce false detections, they cannot detect most markers in the scene. 

Overall, it seems that detecting and decoding markers under complex patterns is indeed more difficult than under uniform lighting; this could stem from the effect of those patterns ``flipping'' some of the bits or at least reducing the certainty of their value (e.g., changing from black to a brighter tone in relation to the surrounding bits). If that is the case, it could be that our methods overly rely on local contrast to detect borders and extract the values of the bits in the marker. Taking this into account, much work could be done to improve detection and decoding by exploiting more high-level details, such as the expected shape of the marker, or even completely disregarding the bit-based approach, by interpreting complete bit ``patterns'' as individual pictographs.

\begin{figure}[t]
    \centering
    \begin{subfigure}[t]{.49\textwidth}
        \centering
        \includegraphics[width=\textwidth]{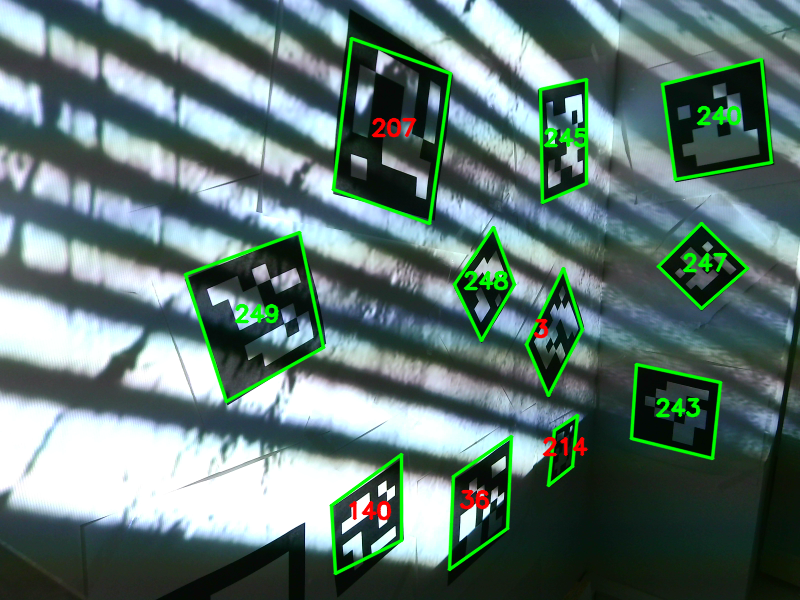}
        \caption{}
    \end{subfigure}
    \begin{subfigure}[t]{.49\textwidth}
        \centering
        \includegraphics[width=\textwidth]{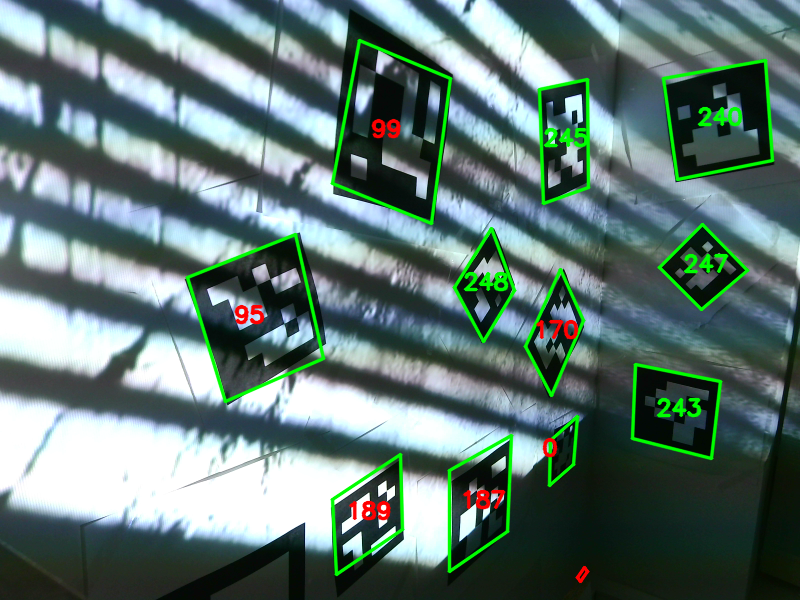}
        \caption{}
    \end{subfigure}
    \caption{\textbf{yolov8s + heatmap vs. yolov8n + direct, on complex lighting settings.} \texttt{yolov8s} + heatmap on (a), \texttt{yolov8n} + direct on (b). Red outlines indicate false positive detections. Red IDs indicate wrong identifications. (Best viewed in digital format)}
    \label{fig:demo_compare}
\end{figure}

In Fig.~\ref{fig:demo_compare}, we can see how the performance of the larger model is not always clearly superior to the smaller one. In this example, \texttt{yolov8s} + heatmap (a) has a similar performance to \texttt{yolov8n} + direct (b), only introducing a few additional errors. Scenarios like these are the most difficult for the method, by combining complex patterns with somewhat extreme marker poses.

%\texttt{yolov8n} + direct (b) manages to recover more markers than \texttt{yolov8s} + heatmap (a), but it introduces additional false detections and marker decoding errors. However, in the light of the results in Table~\ref{tab:sota_comparison} this result may well be anecdotical.

\begin{figure}[t]
    \centering
    \begin{subfigure}[t]{\textwidth}
        \includegraphics[width=.49\textwidth]{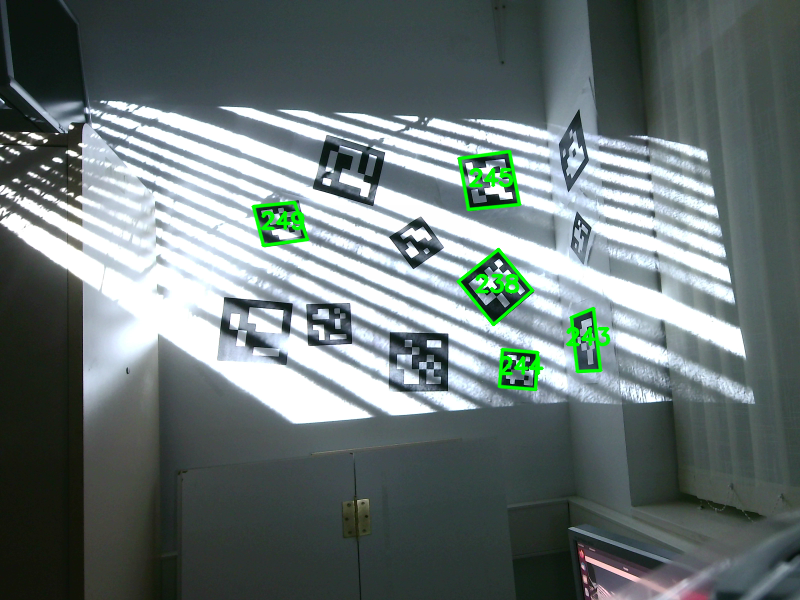}
        \hfill
        \includegraphics[width=.49\textwidth]{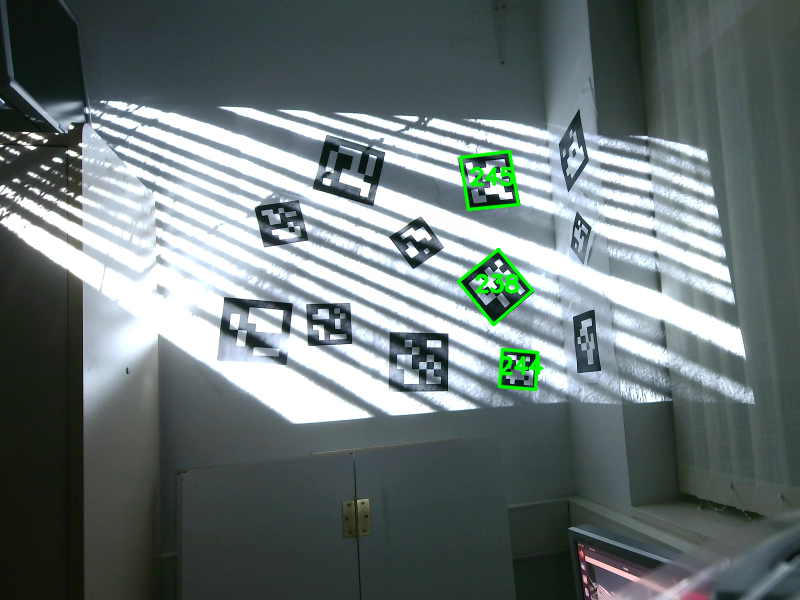}
        \caption{}
        \vspace{1.5mm}
    \end{subfigure}
    \begin{subfigure}[t]{\textwidth}
        \begin{subfigure}[t]{.56\textwidth}
            \includegraphics[height=2.5cm]{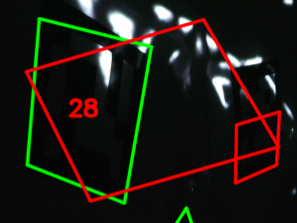}
            \hfill
            \includegraphics[height=2.5cm]{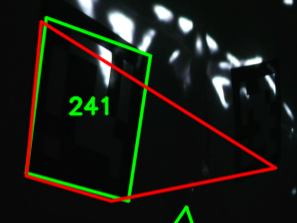}
            \caption{}
        \end{subfigure}
        \hfill
        \begin{subfigure}[t]{.42\textwidth}
            \includegraphics[height=2.5cm]{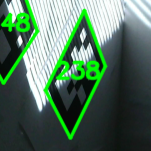}
            \hfill
            \includegraphics[height=2.5cm]{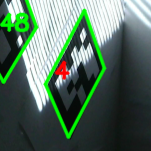}
            \caption{}
        \end{subfigure}
    \end{subfigure}
    \caption{\textbf{Failure cases for multiple DeepArUco++ configurations.} The comparison between the multiple pairs of outputs in this figure highlights many deficiencies and failure cases that can be observed when using our methods. (a) A larger model can detect fewer markers than a smaller one: \texttt{yolov8n} + heatmaps (left) vs. \texttt{yolov8s} + heatmaps (right). (b) Models sometimes hallucinate in pitch-black conditions: \texttt{yolov8n} + direct (left) vs. \texttt{yolov8n} + heatmaps (right). (c) Small variations in predicted corners lead to wrong identifications: \texttt{yolov8n} + direct (left) vs. \texttt{yolov8n} + heatmaps (right). Red outlines indicate false positive detections and red IDs indicate incorrect identifications. Please refer to the main text for an in-depth explanation of the observed situations. (Best viewed in digital format)}
    \label{fig:failures}
\end{figure}

Finally, in Fig.~\ref{fig:failures}, we can see some additional, interesting failure cases that can be found when processing images with our methods. We could highlight the following behaviours: on the one hand, the configurations based on the larger detector (\texttt{yolov8s}) can be somewhat overzealous in scenarios with frequent contrasts between lights and shadows, discarding valid detections. This can be seen on Fig.~\ref{fig:failures}a, when comparing the output from \texttt{yolov8n} + heatmaps (left) against the output from \texttt{yolov8s} + heatmaps (right). The explanation for this would probably stem from a slight overfitting to the training dataset, combined with an imperfect match between the conditions in our synthetic dataset and those in the test dataset. Additionally, our methods may hallucinate in pitch-black conditions, detecting markers where there are none (see Fig.~\ref{fig:failures}b); in this case, the heatmap-based corner refinement model may lead to fewer false detections by not being able to find any corner in the image (right of Fig.~\ref{fig:failures}b), while no detections will be discarded when using the direct regression approach (left of Fig.~\ref{fig:failures}b). Finally, we can see how slight, nearly imperceptible variations in the cropping of a marker will lead to misidentifications: this can be seen in Fig.~\ref{fig:failures}c when comparing the output for \texttt{yolov8n} + direct (left) against the output for \texttt{yolov8n} + heatmaps (right); additional work is needed towards getting a more predictable behaviour from our marker decoder.

%\iffalse
%\begin{figure}[t]
%    \centering
%    \begin{subfigure}[t]{.49\textwidth}
%        \centering
%        \includegraphics[width=\textwidth]{figures/00803_luma_bc_n+reg_hmap_8.png}
%        \caption{}
%    \end{subfigure}
%    \begin{subfigure}[t]{.49\textwidth}
%        \centering
%        \includegraphics[width=\textwidth]{figures/00803_luma_bc_s+reg_hmap_8.png}
%        \caption{}
%    \end{subfigure}
%    \begin{subfigure}[t]{.2742\textwidth}
%        \centering
%        \includegraphics[width=\textwidth]{figures/00836_luma_bc_n+reg_corners_detail.png}
%        \caption{}
%    \end{subfigure}
%    \begin{subfigure}[t]{.2742\textwidth}
%        \centering
%        \includegraphics[width=\textwidth]{figures/00836_luma_bc_n+reg_hmap_8_detail.png}
%        \caption{}
%    \end{subfigure}
%    \begin{subfigure}[t]{.2057\textwidth}
%        \centering
%        \includegraphics[width=\textwidth]{figures/00907_luma_bc_n+reg_corners_detail.png}
%        \caption{}
%    \end{subfigure}
%    \begin{subfigure}[t]{.2057\textwidth}
%        \centering
%        \includegraphics[width=\textwidth]{figures/00907_luma_bc_n+reg_hmap_8_detail.png}
%        \caption{}
%    \end{subfigure}
%    \caption{\new{\textbf{Some failure cases of DeepArUco++ configurations.} \texttt{yolov8n} + heatmaps on (a), (d) and (f); \texttt{yolov8s} + heatmaps on (b); \texttt{yolov8n} + direct on (c) and (e). Red outlines indicate false positive detections. Red IDs indicate wrong identifications. (Best viewed in digital format)}}
    %\label{fig:failures}
%\end{figure}
%\fi

\subsection{Discussion of the Results.}

From the results obtained from the experimentation in Sections~\ref{ssec:sota_comparison} to \ref{ssec:qualresults}, we can draw the following insights.

\textbf{Impact of detection model}. The choice of marker detection model appears to have a relatively minor impact compared to the corner refinement approach. As shown in  Tables~\ref{tab:sota_comparison} and \ref{tab:sota_comparison_deeptag}, the most notable differences arise from using different corner refinement models rather than different detector models. 
This suggests that finding markers in an image is a much simpler task than extracting the precise location of their corners. During the corner refinement step, the direct regression approach is not that useful: while it will always return $4$ predicted corners, they are often placed inaccurately, resulting in imprecise localization. While the error for the remaining corners after filtering is slightly lower with this approach, the heatmap-based approach consistently places more corners within an acceptable distance of their expected locations. Consequently, more markers survive the filtering step with the heatmap-based method.

\textbf{Performance on DeepTag dataset}. On the DeepTag dataset (see Table~\ref{tab:sota_comparison_deeptag}), some trends reverse. During the marker detection step, the configuration using \texttt{yolov8n} as the detector model and direct regression in the corner refinement step correctly decodes more markers than the one using \texttt{yolov8n} and heatmap-based corner refinement. 
This discrepancy may be due to the higher corner error rates observed in the DeepTag dataset compared to the Shadow-ArUco dataset. However, these results might not fully reflect real-world performance due to certain irregularities in the DeepTag dataset, such as the use of the same marker in every image and the presence of ``blockiness'' in the images (see Fig.~\ref{fig:deeptag_detail}). Additionally, the dataset achieves perfect precision in marker decoding, which could indicate some degree of overfitting.

\textbf{Throughput comparison}. In terms of throughput (see Table~\ref{tab:time_breakdown}), the classic ArUco implementation remains the best option, especially in environments with consistent and well-known lighting conditions. Our methods rank second, particularly when using the direct regression approach for the corner refinement step. The loss in throughput with the heatmap-based approach is mainly due to inefficiencies when extracting corner locations from heatmaps. 
DeepTag performs the worst, exhibiting lower average frames per second (FPS) and high variability in frame times, which makes it unsuitable for real-time applications.

\textbf{Qualitative comparison}. From a qualitative perspective, there is ample room for improvement in scenarios with frequent contrasts of light and shadows (see Fig.~\ref{fig:demo_stripes}, \ref{fig:demo_compare} and \ref{fig:failures}a). Our method misses markers or fails to decode them correctly under these conditions. Accurately simulating the variability of real-world lighting in a synthetic dataset is challenging, and enhancing our understanding of the deployment environment could improve performance. Comparatively, detecting markers in poor but homogeneous light is a simpler task (see Fig.~\ref{fig:demo_dark}). Additionally, much work is needed towards reducing hallucinations in pitch-black environments (see Fig.~\ref{fig:failures}b) and improving the consistency in the decoding step (see Fig.~\ref{fig:failures}c).
\section{Conclusions and Future Work} \label{sec:conclusions}

In this work, we have presented a methodology for detecting and decoding square fiducial markers in challenging lighting conditions. To do so, we have defined a pipeline composed of three steps: marker detection, corner refinement, and marker decoding, each based on a different neural network. The described approach is fully modular, allowing different combinations of models to get a configuration better suited for a given task.

To develop our methods, we have created two datasets: the first, fully synthetic (FlyingArUco-v2 dataset), has been used for training the different methods, and the second, obtained by recording physical markers under varying lighting settings (Shadow-ArUco dataset), has been used for testing and comparison against other state-of-the-art techniques. Additionally, we have used a third dataset (DeepTag) to provide a fair comparison outside our datasets.

Our method can outperform classic techniques such as classic ArUco in this task or even other state-of-the-art neural network-based techniques such as DeepTag. Additionally, our method can generalize to other domains outside our target task (e.g., yielding competitive results on the DeepTag dataset, even when compared against the DeepTag method).

As a future work, the method we used to obtain the different models in this work could be extended to a broader range of situations, focusing on additional conditions outside of lighting settings; combining in-the-wild data with our synthetic dataset could also lead to a more robust method. Also, the impact of marker size, distance and angle with respect to the camera, and the presence of blur should be assessed. Finally, further optimization is needed on the implementation side to reduce the hardware requirements to deploy our methods, enabling competition against classic techniques in real-world applications.
\section*{Statements and Declarations}
\begin{itemize}
    \item \textbf{Funding:} Supported by the MCIN Project TED2021-129151B-I00/AEI/ 10.13039/ 501100011033/European Union NextGenerationEU/PRTR, and project PID2023-147296NB-I00 of the Spanish Ministry of Economy, Industry and Competitiveness, FEDER.
    \item \textbf{Competing interests:} The authors have no relevant financial or non-financial interests to disclose.
    \item \textbf{Ethics approval:} For this article, the authors did not undertake work that involved humans or animals.
    \item \textbf{Authors' contributions:} All authors contributed to the study conception and design. Material preparation, data collection, and analysis were performed by Rafael Berral-Soler. All authors commented on previous versions of the manuscript. All authors read and approved the final manuscript.
\end{itemize}

% Bibliography

\bibliographystyle{elsarticle-num}
\bibliography{local}

\end{document}